\documentclass[journal]{IEEEtran}

\usepackage{times}

\usepackage{xcolor,soul,framed} 
\usepackage{threeparttable}
\colorlet{shadecolor}{yellow}
\usepackage[pdftex]{graphicx}
\graphicspath{{../pdf/}{../jpeg/}}
\DeclareGraphicsExtensions{.pdf,.jpeg,.png}

\usepackage[cmex10]{amsmath}
\usepackage{array}
\usepackage{mdwmath}
\usepackage{mdwtab}
\usepackage{eqparbox}
\usepackage{url}
\usepackage{multirow}

\usepackage{times}
\usepackage{helvet}
\usepackage{courier}
\usepackage{hyperref}       
\usepackage{url}   
\usepackage{xcolor}
\usepackage{soul}
\usepackage[utf8]{inputenc}
\usepackage[small]{caption}
\usepackage[cmex10]{amsmath}
\usepackage{epsfig,amsfonts,amssymb,times,subfigure,floatflt,wrapfig,enumerate,tikz,url,bm}

\usepackage{booktabs}
\usepackage{siunitx}
\usepackage{tabu}
\usepackage{multirow}

\newtheorem{theorem}{Theorem}

\newtheorem{definition}{Definition}

\newcommand{\mat}[1]{\bm{#1}}
\newcommand{\ten}[1]{\bm{\mathcal{#1}}}

\usepackage{times}
\usepackage{xcolor}
\usepackage{soul}
\usepackage[utf8]{inputenc}

\usepackage{graphics,epsfig,amsmath,amsfonts,amssymb,times,colortbl,subfigure,floatflt,wrapfig,enumerate,tikz,url,xcolor}
\usepackage{multirow}
\usepackage{eqparbox,comment}
\usepackage{bm}
\usepackage{booktabs}
\usepackage[symbol]{footmisc}
\usepackage{algorithm,algorithmic}

\begin{document}
\bstctlcite{IEEEexample:BSTcontrol}
    \title{Kernelized Support Tensor Train Machines}
  \author{Cong Chen,
      Kim Batselier,
      Wenjian Yu, Senior Member, IEEE, Ngai Wong, Senior Member, IEEE

  \thanks{This work is partially supported by the Hong Kong Research Grants Council under Project 17246416, the University Research Committee of The University of Hong Kong, Tsinghua University Initiative Scientific Research Program, and NSFC under grant No. 61872206.
  
  Cong Chen is with the Department of Electrical and Electronic Engineering,
The University of Hong Kong, Hong Kong. Email: chencong@eee.hku.hk.

Kim Batselier is with the Delft Center for Systems and Control, Delft
University of Technology, Delft, Netherlands. Email: k.batselier@tudelft.nl.

Wenjian Yu is with BNRist, Department of Computer Science
and Technology, Tsinghua University, Beijing 100084, China. Email: yu-wj@tsinghua.edu.cn.

Ngai Wong is with the Department of Electrical and Electronic Engineering,
The University of Hong Kong, Hong Kong. Email: nwong@eee.hku.hk.
  }%
  
  }



\maketitle

\begin{abstract}
Tensor, a multi-dimensional data structure, has been exploited recently in the machine learning community. Traditional machine learning approaches are vector- or matrix-based, and cannot handle tensorial data directly. In this paper, we propose a tensor train (TT)-based kernel technique for the first time, and apply it to the conventional support vector machine (SVM) for image classification. Specifically, we propose a kernelized support tensor train machine that accepts tensorial input and preserves the intrinsic kernel property. The main contributions are threefold. First, we propose a TT-based feature mapping procedure that maintains the TT structure in the feature space. Second, we demonstrate two ways to construct the TT-based kernel function while considering consistency with the TT inner product and preservation of information. Third, we show that it is possible to apply different kernel functions on different data modes. In principle, our method tensorizes the standard SVM on its input structure and kernel mapping scheme. Extensive experiments are performed on real-world tensor data, which demonstrates the superiority of the proposed scheme under few-sample high-dimensional inputs. 
\end{abstract}

\section{INTRODUCTION}
Many real-world data appear in matrix or tensor format. For example, a grayscale picture is a 2-way tensor (i.e. a matrix), a color image or a grayscale video is naturally a 3-way tensor, and a color video can be regarded as a 4-way tensor. In such circumstances, extending the vector-based machine learning algorithms to their tensorial format has recently attracted significant interest in the machine learning and data mining communities.
For example, neighborhood preserving embedding (NPE) was extended to tensor neighborhood preserving embedding (TNPE) in~\cite{dai2006tensor}, principal component analysis to multilinear principal component analysis (MPCA) in~\cite{MPCA}, support vector machines (SVMs)~\cite{boser1992training} to support tensor machines (STMs) in~\cite{STM}, and restricted Boltzmann machines to their tensorial formats in~\cite{nguyen2015tensor}. 

By reformulating the aforementioned machine learning algorithms into the tensorial framework, a huge performance improvement has been achieved. The main reasons for this improvement can be summarized as follows. Firstly, these tensorized algorithms can naturally utilize the multi-way structure of the original tensor data, which is believed to be useful in many machine learning applications such as pattern recognition~\cite{phan2010tensor}, image completion~\cite{liu2013tensor} and anomaly detection~\cite{fanaee2016tensor}. 
Secondly, vectorizing tensor data leads to  high-dimensional vectors, which may cause overfitting especially when the training sample size is relatively small~\cite{guo2012tensor}. On the contrary, tensor-based approaches usually derive a more structural and robust model that commonly involves much fewer model parameters, which not only alleviates the overfitting problem, but also saves a lot of storage and computation resources~\cite{lebedev2014speeding,novikov2015tensorizing}. 

In this paper, we propose a kernelized support tensor train machine (K-STTM) to address few-sample image classification problems due to the fact that collecting labeled pictures is very expensive and time-consuming in many research areas. Specifically, we first employ the tensor train (TT) decomposition~\cite{oseledets2011tensor} to decompose the given tensor data so that a more compact and informative representation of it can be derived. Secondly, we define a TT-based feature mapping strategy to derive a high-dimensional TT in the feature space. This strategy enables us to apply different feature mappings on different data modes, which naturally provides a way to leverage the multi-modal nature of tensor structured data. Thirdly, we propose two ways to build the kernel matrix with the consideration of the consistency with the TT inner product and preservation of information. The constructed kernel matrix is then used by kernel machines to solve the image classification problems.

There are two main advantages from the proposed methods. On the one hand, the proposed methods are naturally nonlinear classifiers. It is common that real-life data are not linearly separable. However, most existing supervised tensor learning methods which employ tensor input are often based on a linear model and cannot deal with nonlinear classification problems. In that case, our proposed methods can handle nonlinear learning problems on tensor data better. On the other hand, conventional tensor-based kernel methods focus on flatting tensor data into matrices~\cite{signoretto2011kernel,zhao2013kernel}, and thus can only preserve one-mode relationships within the tensor itself. However, our proposed approaches can capture and exploit multi-mode relationships, which commonly leads to more powerful and accurate models.  

The superiority of our methods is validated through extensive experiments. It is observed that our methods achieve a much better performance than the linear supervised tensor learning methods, which indicates the importance of introducing kernel trick. Furthermore, our methods achieve a better classification performance when the input data are truly high-dimensional. Applying different kernel functions on different data modes is also investigated and shows an obvious improvement compared with the baseline.

The rest of this paper is organized as follows. In Section~\ref{sec:relatedworks}, we briefly review some related works in supervised tensor learning. Some useful notations and tensor arithmetic are further introduced in Section~\ref{sec:prelim}. In Section~\ref{sec:ksttm}, we formulate the proposed kernelized support tensor train machine (K-STTM). Experiments are shown in Section~\ref{sec:experiments} to validate the superiority of our methods. 
Lastly, we draw conclusions and propose some possible extended works in Section~\ref{sec:conclusion}.

\section{RELATED WORKS}
\label{sec:relatedworks}
As one of the most typical supervised learning algorithms, SVM~\cite{boser1992training} has achieved an enormous success in pattern classification by minimizing the Vapnik-Chervonenkis dimensions and structural risk. However, a standard SVM can not deal with tensorial input directly. The first work that extends SVM to handle tensorial input is~\cite{STM}. More precisely, a supervised tensor learning (STL) scheme was proposed to train a support tensor machine (STM), where the hyperplane parameters are modeled as a rank-1 tensor instead of a vector. For the parameter training, they employed the alternating projection optimization  method.  

Although STM is capable to classify tensorial data directly, the  expressive power of a rank-1 weight tensor is limited, which often leads to a poor classification accuracy. To increase the model expression capacity, several works were proposed recently based on the STL scheme. Ref.~\cite{kotsia2012higher} employs a more general tensor structure, i.e., the canonical polyadic (CP) format, to replace the rank-1 weight tensor in STM. However, it is an NP-complete problem to determine the CP-rank. In~\cite{kotsia2011support}, the STM is generalized to a support Tucker machine (STuM) by representing the weight parameter as a Tucker tensor. Nevertheless, the number of model parameters in STuM is exponentially large, which often leads to a large amount of storage and computation consumption. To overcome this, Ref.~\cite{linearSTTM} proposed a support tensor train machine (STTM), which assumes the potential weight tensor format is a TT. By doing so, the corresponding optimization problem is more scalable and can be solved efficiently. The aforementioned work are all based on the assumption that the given tensorial data are linearly separable. However, this is not the case in most real-world data. It is worth noting that though STTM sounds like the linear case of the proposed K-STTM, they are totally different when the linear kernel is applied on K-STTM. Specifically, K-STTM and STTM use two totally different schemes to train the corresponding model. For K-STTM, it first constructs the kernel matrix with the proposed TT-based kernel function, and then solves the standard SVM problem. However, in STTM, it assumes the parameter in the classification hyperplane can be modeled as a TT, and only updates one TT-core at a time by reformulating the training data.

To extend the linear tensorial classifiers to the nonlinear case, the authors in~\cite{he2014dusk} proposed a nonlinear supervised learning scheme called dual structure-preserving kernels (DuSK). Specifically, based on the CP tensor structure, they define a corresponding kernel trick to map the CP format data into a higher-dimensional feature space. Through the introduction of the kernel trick, DuSK can achieve a higher classification accuracy. However, since DuSK is based on the CP decomposition, the NP-complete problem on the rank determination still exists. Moreover, through introducing a kernelized CP tensor factorization technique, the same research group in~\cite{he2014dusk} further proposed the Multi-way
Multi-level Kernel model~\cite{he2017multi} and kernelized support tensor machine model~\cite{he2017kernelized}. Nevertheless, the CP-rank determination issue still exists since they are all based on the CP decomposition.

To avoid the above issues, we propose the K-STTM, which not only introduces the customized kernel function to handle nonlinear classification problems, but also achieves an efficient model training since the scalable TT format is employed.    
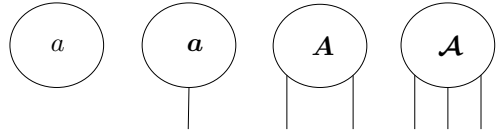
\begin{figure}[tb] 
\begin{center} 
\ifx\du\undefined
  \newlength{\du}
\fi
\setlength{\du}{15\unitlength}
\begin{tikzpicture}
\pgftransformxscale{0.750000}
\pgftransformyscale{-0.700000}
\definecolor{dialinecolor}{rgb}{0.000000, 0.000000, 0.000000}
\pgfsetstrokecolor{dialinecolor}
\definecolor{dialinecolor}{rgb}{1.000000, 1.000000, 1.000000}
\pgfsetfillcolor{dialinecolor}
\pgfsetlinewidth{0.00000\du}
\pgfsetdash{}{0pt}
\pgfsetdash{}{0pt}
\pgfsetmiterjoin
\definecolor{dialinecolor}{rgb}{0.000000, 0.000000, 0.000000}
\pgfsetstrokecolor{dialinecolor}
\pgfpathellipse{\pgfpoint{13.375000\du}{11.975000\du}}{\pgfpoint{1.575000\du}{0\du}}{\pgfpoint{0\du}{1.525000\du}}
\pgfusepath{stroke}
\definecolor{dialinecolor}{rgb}{0.000000, 0.000000, 0.000000}
\pgfsetstrokecolor{dialinecolor}
\node at (13.375000\du,12.215000\du){};
\pgfsetlinewidth{0.00000\du}
\pgfsetdash{}{0pt}
\pgfsetdash{}{0pt}
\pgfsetbuttcap
{
\definecolor{dialinecolor}{rgb}{0.000000, 0.000000, 0.000000}
\pgfsetfillcolor{dialinecolor}
\definecolor{dialinecolor}{rgb}{0.000000, 0.000000, 0.000000}
\pgfsetstrokecolor{dialinecolor}
\draw (13.375000\du,13.500000\du)--(13.350000\du,15.0000\du);
}
\pgfsetlinewidth{0.00000\du}
\pgfsetdash{}{0pt}
\pgfsetdash{}{0pt}
\pgfsetmiterjoin
\definecolor{dialinecolor}{rgb}{0.000000, 0.000000, 0.000000}
\pgfsetstrokecolor{dialinecolor}
\pgfpathellipse{\pgfpoint{8.930000\du}{11.975000\du}}{\pgfpoint{1.575000\du}{0\du}}{\pgfpoint{0\du}{1.525000\du}}
\pgfusepath{stroke}
\definecolor{dialinecolor}{rgb}{0.000000, 0.000000, 0.000000}
\pgfsetstrokecolor{dialinecolor}
\node at (8.930000\du,12.215000\du){};
\pgfsetlinewidth{0.00000\du}
\pgfsetdash{}{0pt}
\pgfsetdash{}{0pt}
\pgfsetmiterjoin
\definecolor{dialinecolor}{rgb}{0.000000, 0.000000, 0.000000}
\pgfsetstrokecolor{dialinecolor}
\pgfpathellipse{\pgfpoint{17.780000\du}{12.000000\du}}{\pgfpoint{1.575000\du}{0\du}}{\pgfpoint{0\du}{1.500000\du}}
\pgfusepath{stroke}
\definecolor{dialinecolor}{rgb}{0.000000, 0.000000, 0.000000}
\pgfsetstrokecolor{dialinecolor}
\node at (17.780000\du,12.240000\du){};
\pgfsetlinewidth{0.00000\du}
\pgfsetdash{}{0pt}
\pgfsetdash{}{0pt}
\pgfsetbuttcap
{
\definecolor{dialinecolor}{rgb}{0.000000, 0.000000, 0.000000}
\pgfsetfillcolor{dialinecolor}
\definecolor{dialinecolor}{rgb}{0.000000, 0.000000, 0.000000}
\pgfsetstrokecolor{dialinecolor}
\draw (18.893693\du,13.060660\du)--(18.893693\du,15.000000\du);
}
\pgfsetlinewidth{0.00000\du}
\pgfsetdash{}{0pt}
\pgfsetdash{}{0pt}
\pgfsetbuttcap
{
\definecolor{dialinecolor}{rgb}{0.000000, 0.000000, 0.000000}
\pgfsetfillcolor{dialinecolor}
\definecolor{dialinecolor}{rgb}{0.000000, 0.000000, 0.000000}
\pgfsetstrokecolor{dialinecolor}
\draw (16.666307\du,13.060660\du)--(16.666307\du,15.00000\du);
}
\pgfsetlinewidth{0.00000\du}
\pgfsetdash{}{0pt}
\pgfsetdash{}{0pt}
\pgfsetmiterjoin
\definecolor{dialinecolor}{rgb}{0.000000, 0.000000, 0.000000}
\pgfsetstrokecolor{dialinecolor}
\pgfpathellipse{\pgfpoint{22.080000\du}{11.975000\du}}{\pgfpoint{1.575000\du}{0\du}}{\pgfpoint{0\du}{1.525000\du}}
\pgfusepath{stroke}
\definecolor{dialinecolor}{rgb}{0.000000, 0.000000, 0.000000}
\pgfsetstrokecolor{dialinecolor}
\node at (22.080000\du,12.215000\du){};
\pgfsetlinewidth{0.00000\du}
\pgfsetdash{}{0pt}
\pgfsetdash{}{0pt}
\pgfsetbuttcap
{
\definecolor{dialinecolor}{rgb}{0.000000, 0.000000, 0.000000}
\pgfsetfillcolor{dialinecolor}
\definecolor{dialinecolor}{rgb}{0.000000, 0.000000, 0.000000}
\pgfsetstrokecolor{dialinecolor}
\draw (23.193693\du,13.053338\du)--(23.193693\du,15.000000\du);
}
\pgfsetlinewidth{0.00000\du}
\pgfsetdash{}{0pt}
\pgfsetdash{}{0pt}
\pgfsetbuttcap
{
\definecolor{dialinecolor}{rgb}{0.000000, 0.000000, 0.000000}
\pgfsetfillcolor{dialinecolor}
\definecolor{dialinecolor}{rgb}{0.000000, 0.000000, 0.000000}
\pgfsetstrokecolor{dialinecolor}
\draw (20.966307\du,13.053338\du)--(20.966307\du,15.000000\du);
}
\pgfsetlinewidth{0.00000\du}
\pgfsetdash{}{0pt}
\pgfsetdash{}{0pt}
\pgfsetbuttcap
{
\definecolor{dialinecolor}{rgb}{0.000000, 0.000000, 0.000000}
\pgfsetfillcolor{dialinecolor}
\definecolor{dialinecolor}{rgb}{0.000000, 0.000000, 0.000000}
\pgfsetstrokecolor{dialinecolor}
\draw (22.080000\du,13.500000\du)--(22.080000\du,15.000000\du);
}
\definecolor{dialinecolor}{rgb}{0.000000, 0.000000, 0.000000}
\pgfsetstrokecolor{dialinecolor}
\node[anchor=west] at (8.430000\du,11.975000\du){$a$};
\definecolor{dialinecolor}{rgb}{0.000000, 0.000000, 0.000000}
\pgfsetstrokecolor{dialinecolor}
\node[anchor=west] at (9.200000\du,11.900000\du){};
\definecolor{dialinecolor}{rgb}{0.000000, 0.000000, 0.000000}
\pgfsetstrokecolor{dialinecolor}
\node[anchor=west] at (13.005000\du,11.975000\du){$\mat{a}$};
\definecolor{dialinecolor}{rgb}{0.000000, 0.000000, 0.000000}
\pgfsetstrokecolor{dialinecolor}
\node[anchor=west] at (17.200000\du,12.000000\du){$\mat{A}$};
\definecolor{dialinecolor}{rgb}{0.000000, 0.000000, 0.000000}
\pgfsetstrokecolor{dialinecolor}
\node[anchor=west] at (21.450000\du,12.00000\du){$\ten{A}$};
\end{tikzpicture}
\caption{Graphical representation of a scalar $a$, vector $\mat{a}$, matrix $\mat{A}$, and third-order tensor $\ten{A}$.}
\label{fig:graphical}
\end{center}
\end{figure}

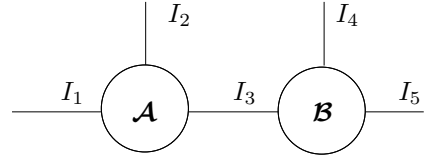
\begin{figure}[tb] 
\begin{center} 
\ifx\du\undefined
  \newlength{\du}
\fi
\setlength{\du}{15\unitlength}
\begin{tikzpicture}
\pgftransformxscale{0.750000}
\pgftransformyscale{-0.750000}
\definecolor{dialinecolor}{rgb}{0.000000, 0.000000, 0.000000}
\pgfsetstrokecolor{dialinecolor}
\definecolor{dialinecolor}{rgb}{1.000000, 1.000000, 1.000000}
\pgfsetfillcolor{dialinecolor}
\pgfsetlinewidth{0.00000\du}
\pgfsetdash{}{0pt}
\pgfsetdash{}{0pt}
\pgfsetbuttcap
\pgfsetmiterjoin
\pgfsetlinewidth{0.00000\du}
\pgfsetbuttcap
\pgfsetmiterjoin
\pgfsetdash{}{0pt}
\definecolor{dialinecolor}{rgb}{1.000000, 1.000000, 1.000000}
\pgfsetfillcolor{dialinecolor}
\pgfpathellipse{\pgfpoint{6.573437\du}{8.535937\du}}{\pgfpoint{1.464062\du}{0\du}}{\pgfpoint{0\du}{1.464062\du}}
\pgfusepath{fill}
\definecolor{dialinecolor}{rgb}{0.000000, 0.000000, 0.000000}
\pgfsetstrokecolor{dialinecolor}
\pgfpathellipse{\pgfpoint{6.573437\du}{8.535937\du}}{\pgfpoint{1.464062\du}{0\du}}{\pgfpoint{0\du}{1.464062\du}}
\pgfusepath{stroke}
\pgfsetbuttcap
\pgfsetmiterjoin
\pgfsetdash{}{0pt}
\definecolor{dialinecolor}{rgb}{0.000000, 0.000000, 0.000000}
\pgfsetstrokecolor{dialinecolor}
\pgfpathellipse{\pgfpoint{6.573437\du}{8.535937\du}}{\pgfpoint{1.464062\du}{0\du}}{\pgfpoint{0\du}{1.464062\du}}
\pgfusepath{stroke}
\pgfsetlinewidth{0.00000\du}
\pgfsetdash{}{0pt}
\pgfsetdash{}{0pt}
\pgfsetbuttcap
\pgfsetmiterjoin
\pgfsetlinewidth{0.00000\du}
\pgfsetbuttcap
\pgfsetmiterjoin
\pgfsetdash{}{0pt}
\definecolor{dialinecolor}{rgb}{1.000000, 1.000000, 1.000000}
\pgfsetfillcolor{dialinecolor}
\pgfpathellipse{\pgfpoint{12.519062\du}{8.609062\du}}{\pgfpoint{1.464062\du}{0\du}}{\pgfpoint{0\du}{1.464062\du}}
\pgfusepath{fill}
\definecolor{dialinecolor}{rgb}{0.000000, 0.000000, 0.000000}
\pgfsetstrokecolor{dialinecolor}
\pgfpathellipse{\pgfpoint{12.519062\du}{8.609062\du}}{\pgfpoint{1.464062\du}{0\du}}{\pgfpoint{0\du}{1.464062\du}}
\pgfusepath{stroke}
\pgfsetbuttcap
\pgfsetmiterjoin
\pgfsetdash{}{0pt}
\definecolor{dialinecolor}{rgb}{0.000000, 0.000000, 0.000000}
\pgfsetstrokecolor{dialinecolor}
\pgfpathellipse{\pgfpoint{12.519062\du}{8.609062\du}}{\pgfpoint{1.464062\du}{0\du}}{\pgfpoint{0\du}{1.464062\du}}
\pgfusepath{stroke}
\pgfsetlinewidth{0.00000\du}
\pgfsetdash{}{0pt}
\pgfsetdash{}{0pt}
\pgfsetbuttcap
{
\definecolor{dialinecolor}{rgb}{0.000000, 0.000000, 0.000000}
\pgfsetfillcolor{dialinecolor}
\definecolor{dialinecolor}{rgb}{0.000000, 0.000000, 0.000000}
\pgfsetstrokecolor{dialinecolor}
\draw (8.037500\du,8.65000\du)--(11.055000\du,8.65000\du);
}
\pgfsetlinewidth{0.00000\du}
\pgfsetdash{}{0pt}
\pgfsetdash{}{0pt}
\pgfsetbuttcap
{
\definecolor{dialinecolor}{rgb}{0.000000, 0.000000, 0.000000}
\pgfsetfillcolor{dialinecolor}
\definecolor{dialinecolor}{rgb}{0.000000, 0.000000, 0.000000}
\pgfsetstrokecolor{dialinecolor}
\draw (2.106197\du,8.65000\du)--(5.109375\du,8.65000\du);
}
\pgfsetlinewidth{0.00000\du}
\pgfsetdash{}{0pt}
\pgfsetdash{}{0pt}
\pgfsetbuttcap
{
\definecolor{dialinecolor}{rgb}{0.000000, 0.000000, 0.000000}
\pgfsetfillcolor{dialinecolor}
\definecolor{dialinecolor}{rgb}{0.000000, 0.000000, 0.000000}
\pgfsetstrokecolor{dialinecolor}
\draw (6.600000\du,7.071875\du)--(6.600000\du,4.950000\du);
}

\pgfsetlinewidth{-0.000\du}
\pgfsetdash{}{0pt}
\pgfsetdash{}{0pt}
\pgfsetbuttcap
{
\definecolor{dialinecolor}{rgb}{0.000000, 0.000000, 0.000000}
\pgfsetfillcolor{dialinecolor}
\definecolor{dialinecolor}{rgb}{0.000000, 0.000000, 0.000000}
\pgfsetstrokecolor{dialinecolor}
\draw (12.600000\du,7.091875\du)--(12.600000\du,4.950000\du);
}

\pgfsetlinewidth{0.00000\du}
\pgfsetdash{}{0pt}
\pgfsetdash{}{0pt}
\pgfsetbuttcap
{
\definecolor{dialinecolor}{rgb}{0.000000, 0.000000, 0.000000}
\pgfsetfillcolor{dialinecolor}
\definecolor{dialinecolor}{rgb}{0.000000, 0.000000, 0.000000}
\pgfsetstrokecolor{dialinecolor}
\draw (13.983125\du,8.65000\du)--(15.950000\du,8.650000\du);
}
\definecolor{dialinecolor}{rgb}{0.000000, 0.000000, 0.000000}
\pgfsetstrokecolor{dialinecolor}
\node[anchor=west] at (3.450000\du,8.000000\du){$I_1$};
\definecolor{dialinecolor}{rgb}{0.000000, 0.000000, 0.000000}
\pgfsetstrokecolor{dialinecolor}
\node[anchor=west] at (7.050000\du,5.400000\du){$I_2$};
\definecolor{dialinecolor}{rgb}{0.000000, 0.000000, 0.000000}
\pgfsetstrokecolor{dialinecolor}
\node[anchor=west] at (12.70000\du,5.400000\du){$I_4$};
\definecolor{dialinecolor}{rgb}{0.000000, 0.000000, 0.000000}
\pgfsetstrokecolor{dialinecolor}
\node[anchor=west] at (9.250000\du,8.000000\du){$I_3$};
\definecolor{dialinecolor}{rgb}{0.000000, 0.000000, 0.000000}
\pgfsetstrokecolor{dialinecolor}
\node[anchor=west] at (14.800000\du,8.000000\du){$I_5$};
\definecolor{dialinecolor}{rgb}{0.000000, 0.000000, 0.000000}
\pgfsetstrokecolor{dialinecolor}
\node[anchor=west] at (5.83437\du,8.65000\du){$\ten{A}$};
\definecolor{dialinecolor}{rgb}{0.000000, 0.000000, 0.000000}
\pgfsetstrokecolor{dialinecolor}
\node[anchor=west] at (11.91062\du,8.65000\du){$\ten{B}$};
\end{tikzpicture}
\caption{Index contraction between two 3-way tensors $\ten{A}$ and $\ten{B}$.}
\label{fig:kmodeprod}
\end{center}
\end{figure}
 
\section{PRELIMINARIES  }
\label{sec:prelim}
In this Section, we review some basic tensor notations and operations, together with the related tensor train decomposition method.
\subsection{Tensor basics}
Tensors in this paper are multi-dimensional arrays that generalize vectors (first-order tensors) and matrices (second-order tensors) to higher orders. A $d$th-order or $d$-way tensor is denoted as $\ten{A}\in\mathbb{R}^{I_1\times I_2 \times \cdots \times I_d}$ and the element of $\ten{A}$ by $\ten{A}(i_1,i_2\ldots,i_d)$, where 1$\le$ $i_k$ $\le$ $I_k$, $k = 1,2,\ldots,d$. The numbers $I_1,  I_2,\ldots, I_d$ are called the dimensions of the tensor $\ten{A}$.
We use boldface capital calligraphic letters $\ten{A}$, $\ten{B}$, \ldots to denote tensors, boldface capital letters $\mat{A}$, $\mat{B}$, \ldots to denote matrices, boldface letters $\mat{a}$, $\mat{b}$, \ldots to denote vectors, and roman letters $a$, $b$, \ldots to denote scalars. An intuitive and useful graphical representation of scalars, vectors, matrices and tensors is depicted in Figure~\ref{fig:graphical}. The unconnected edges, also called free legs, are the indices of the tensor. Therefore scalars have no free legs, while a matrix has 2 free legs. We will employ these graphical representations to visualize the tensor networks and operations in the following sections whenever possible and refer to~\cite{orus2014practical} for more details. We now briefly introduce some important tensor operations.
\begin{definition}(Tensor index contraction): A tensor index contraction is the sum over all possible values of the repeated indices in a set of tensors. 
\end{definition}
For example, the following contraction of two 3-way tensors $\ten{A}$ and $\ten{B}$ 
\begin{align}
 \ten{C}(i_1,i_2,i_4,i_5)&=
\nonumber  \sum\limits_{i_3=1}^{I_3}  \ten{A}(i_1,i_2,i_3)\, \ten{B}(i_3,i_4,i_5),
\end{align}
over the $i_3$ index produces a four-way tensor $\ten{C}$. We also present the graphical representation of this contraction in Figure~\ref{fig:kmodeprod}, where the summation over $i_3$ is indicated by the connected edge. After this contraction, the tensor diagram contains four free legs indexed by $i_1,i_2,i_4,i_5$, respectively.

\begin{definition}(Tensor inner product):
For two tensors $\ten{A},\ten{B} \in \mathbb{R}^{I_1 \times I_2 \times \cdots \times I_d}$, their inner product $\langle \ten{A},\ten{B} \rangle$ is defined as
\begin{align*}
\langle \ten{A},\ten{B} \rangle &= \sum\limits_{i_1=1}^{I_1} \sum\limits_{i_2=1}^{I_2}\cdots \sum\limits_{i_d=1}^{I_d} a_{i_1,i_2,\cdots,i_d} b_{i_1,i_2,\cdots,i_d}.
\end{align*}
\end{definition}

\begin{definition}(Tensor Frobenius norm): The Frobenius norm of a tensor $\ten{A}\in\mathbb{R}^{I_1\times I_2 \times \cdots\times I_d}$ is defined as $||\ten{A}||_F=\sqrt{\langle \ten{A},\ten{A}\rangle}$.
\end{definition}

\subsection{Tensor train decomposition}
\label{subsec:TT}
\begin{figure}[t]
\begin{center}
\ifx\du\undefined
  \newlength{\du}
\fi
\setlength{\du}{15\unitlength}
\begin{tikzpicture}
\pgftransformxscale{1.000000}
\pgftransformyscale{-1.000000}
\definecolor{dialinecolor}{rgb}{0.000000, 0.000000, 0.000000}
\pgfsetstrokecolor{dialinecolor}
\definecolor{dialinecolor}{rgb}{1.000000, 1.000000, 1.000000}
\pgfsetfillcolor{dialinecolor}
\pgfsetlinewidth{0.000000\du}
\pgfsetdash{}{0pt}
\pgfsetdash{}{0pt}
\pgfsetbuttcap
\pgfsetmiterjoin
\pgfsetlinewidth{0.000000\du}
\pgfsetbuttcap
\pgfsetmiterjoin
\pgfsetdash{}{0pt}
\definecolor{dialinecolor}{rgb}{1.000000, 1.000000, 1.000000}
\pgfsetfillcolor{dialinecolor}
\pgfpathellipse{\pgfpoint{7.8\du}{5.500000\du}}{\pgfpoint{1.000000\du}{0\du}}{\pgfpoint{0\du}{1.000000\du}}
\pgfusepath{fill}
\definecolor{dialinecolor}{rgb}{0.000000, 0.000000, 0.000000}
\pgfsetstrokecolor{dialinecolor}
\pgfpathellipse{\pgfpoint{7.8\du}{5.500000\du}}{\pgfpoint{1.000000\du}{0\du}}{\pgfpoint{0\du}{1.000000\du}}
\pgfusepath{stroke}
\pgfsetbuttcap
\pgfsetmiterjoin
\pgfsetdash{}{0pt}
\definecolor{dialinecolor}{rgb}{0.000000, 0.000000, 0.000000}
\pgfsetstrokecolor{dialinecolor}
\pgfpathellipse{\pgfpoint{7.8\du}{5.500000\du}}{\pgfpoint{1.000000\du}{0\du}}{\pgfpoint{0\du}{1.000000\du}}
\pgfusepath{stroke}
\pgfsetlinewidth{0.000000\du}
\pgfsetdash{}{0pt}
\pgfsetdash{}{0pt}
\pgfsetbuttcap
\pgfsetmiterjoin
\pgfsetlinewidth{0.00000\du}
\pgfsetbuttcap
\pgfsetmiterjoin
\pgfsetdash{}{0pt}
\definecolor{dialinecolor}{rgb}{1.000000, 1.000000, 1.000000}
\pgfsetfillcolor{dialinecolor}
\pgfpathellipse{\pgfpoint{11.2\du}{5.500000\du}}{\pgfpoint{1.000000\du}{0\du}}{\pgfpoint{0\du}{1.000000\du}}
\pgfusepath{fill}
\definecolor{dialinecolor}{rgb}{0.000000, 0.000000, 0.000000}
\pgfsetstrokecolor{dialinecolor}
\pgfpathellipse{\pgfpoint{11.2\du}{5.50000\du}}{\pgfpoint{1.000000\du}{0\du}}{\pgfpoint{0\du}{1.000000\du}}
\pgfusepath{stroke}
\pgfsetbuttcap
\pgfsetmiterjoin
\pgfsetdash{}{0pt}
\definecolor{dialinecolor}{rgb}{0.000000, 0.000000, 0.000000}
\pgfsetstrokecolor{dialinecolor}
\pgfpathellipse{\pgfpoint{11.2\du}{5.50000\du}}{\pgfpoint{1.000000\du}{0\du}}{\pgfpoint{0\du}{1.000000\du}}
\pgfusepath{stroke}
\pgfsetlinewidth{0.00000\du}
\pgfsetdash{}{0pt}
\pgfsetdash{}{0pt}
\pgfsetbuttcap
\pgfsetmiterjoin
\pgfsetlinewidth{0.00000\du}
\pgfsetbuttcap
\pgfsetmiterjoin
\pgfsetdash{}{0pt}
\definecolor{dialinecolor}{rgb}{1.000000, 1.000000, 1.000000}
\pgfsetfillcolor{dialinecolor}
\pgfpathellipse{\pgfpoint{17.3\du}{5.500000\du}}{\pgfpoint{1.000000\du}{0\du}}{\pgfpoint{0\du}{1.000000\du}}
\pgfusepath{fill}
\definecolor{dialinecolor}{rgb}{0.000000, 0.000000, 0.000000}
\pgfsetstrokecolor{dialinecolor}
\pgfpathellipse{\pgfpoint{17.3\du}{5.500000\du}}{\pgfpoint{1.000000\du}{0\du}}{\pgfpoint{0\du}{1.000000\du}}
\pgfusepath{stroke}
\pgfsetbuttcap
\pgfsetmiterjoin
\pgfsetdash{}{0pt}
\definecolor{dialinecolor}{rgb}{0.000000, 0.000000, 0.000000}
\pgfsetstrokecolor{dialinecolor}
\pgfpathellipse{\pgfpoint{17.3\du}{5.500000\du}}{\pgfpoint{1.000000\du}{0\du}}{\pgfpoint{0\du}{1.000000\du}}
\pgfusepath{stroke}
\pgfsetlinewidth{0.00000\du}
\pgfsetdash{}{0pt}
\pgfsetdash{}{0pt}
\pgfsetbuttcap
{
\definecolor{dialinecolor}{rgb}{0.000000, 0.000000, 0.000000}
\pgfsetfillcolor{dialinecolor}
\definecolor{dialinecolor}{rgb}{0.000000, 0.000000, 0.000000}
\pgfsetstrokecolor{dialinecolor}
\draw (8.8\du,5.5\du)--(10.2\du,5.5\du);
}
\pgfsetlinewidth{0.00000\du}
\pgfsetdash{}{0pt}
\pgfsetdash{}{0pt}
\pgfsetbuttcap
{
\definecolor{dialinecolor}{rgb}{0.000000, 0.000000, 0.000000}
\pgfsetfillcolor{dialinecolor}
\definecolor{dialinecolor}{rgb}{0.000000, 0.000000, 0.000000}
\pgfsetstrokecolor{dialinecolor}
\draw (12.2\du,5.50000\du)--(13.100000\du,5.500000\du);
}
\pgfsetlinewidth{0.00000\du}
\pgfsetdash{}{0pt}
\pgfsetdash{}{0pt}
\pgfsetbuttcap
{
\definecolor{dialinecolor}{rgb}{0.000000, 0.000000, 0.000000}
\pgfsetfillcolor{dialinecolor}
\definecolor{dialinecolor}{rgb}{0.000000, 0.000000, 0.000000}
\pgfsetstrokecolor{dialinecolor}
\draw (14.8\du,5.5\du)--(16.3\du,5.500000\du);
}
\pgfsetlinewidth{0.00000\du}
\pgfsetdash{}{0pt}
\pgfsetdash{}{0pt}
\pgfsetbuttcap
{
\definecolor{dialinecolor}{rgb}{0.000000, 0.000000, 0.000000}
\pgfsetfillcolor{dialinecolor}
\definecolor{dialinecolor}{rgb}{0.000000, 0.000000, 0.000000}
\pgfsetstrokecolor{dialinecolor}
\draw (7.8\du,6.5\du)--(7.8\du,7.60000\du);
}
\pgfsetlinewidth{0.00000\du}
\pgfsetdash{}{0pt}
\pgfsetdash{}{0pt}
\pgfsetbuttcap
{
\definecolor{dialinecolor}{rgb}{0.000000, 0.000000, 0.000000}
\pgfsetfillcolor{dialinecolor}
\definecolor{dialinecolor}{rgb}{0.000000, 0.000000, 0.000000}
\pgfsetstrokecolor{dialinecolor}
\draw (11.2\du,6.50000\du)--(11.2\du,7.60000\du);
}
\pgfsetlinewidth{0.00000\du}
\pgfsetdash{}{0pt}
\pgfsetdash{}{0pt}
\pgfsetbuttcap
{
\definecolor{dialinecolor}{rgb}{0.000000, 0.000000, 0.000000}
\pgfsetfillcolor{dialinecolor}
\definecolor{dialinecolor}{rgb}{0.000000, 0.000000, 0.000000}
\pgfsetstrokecolor{dialinecolor}
\draw (17.25\du,6.600000\du)--(17.250000\du,7.60000\du);
}
\definecolor{dialinecolor}{rgb}{0.000000, 0.000000, 0.000000}
\pgfsetstrokecolor{dialinecolor}
\node[anchor=west] at (7.2\du,5.50000\du){$\ten{A}^{(1)}$};
\definecolor{dialinecolor}{rgb}{0.000000, 0.000000, 0.000000}
\pgfsetstrokecolor{dialinecolor}
\node[anchor=west] at (10.61816\du,5.50000\du){$\ten{A}^{(2)}$};
\definecolor{dialinecolor}{rgb}{0.000000, 0.000000, 0.000000}
\pgfsetstrokecolor{dialinecolor}
\node[anchor=west] at (16.606816\du,5.50000\du){$\ten{A}^{(d)}$};
\definecolor{dialinecolor}{rgb}{0.000000, 0.000000, 0.000000}
\pgfsetstrokecolor{dialinecolor}
\node[anchor=west] at (13.5\du,5.50000\du){...};
\pgfsetlinewidth{0.00000\du}
\pgfsetdash{}{0pt}
\pgfsetdash{}{0pt}
\pgfsetbuttcap
{
\definecolor{dialinecolor}{rgb}{0.000000, 0.000000, 0.000000}
\pgfsetfillcolor{dialinecolor}
\definecolor{dialinecolor}{rgb}{0.000000, 0.000000, 0.000000}
\pgfsetstrokecolor{dialinecolor}
\draw (18.3\du,5.500000\du)--(19.7\du,5.5\du);
}
\pgfsetlinewidth{0.00000\du}
\pgfsetdash{}{0pt}
\pgfsetdash{}{0pt}
\pgfsetbuttcap
{
\definecolor{dialinecolor}{rgb}{0.000000, 0.000000, 0.000000}
\pgfsetfillcolor{dialinecolor}
\definecolor{dialinecolor}{rgb}{0.000000, 0.000000, 0.000000}
\pgfsetstrokecolor{dialinecolor}
\draw (5.4\du,5.50000\du)--(6.8\du,5.50000\du);
}
\definecolor{dialinecolor}{rgb}{0.000000, 0.000000, 0.000000}
\pgfsetstrokecolor{dialinecolor}
\node[anchor=west] at (5.5\du,5.10000\du){$R_1$};
\definecolor{dialinecolor}{rgb}{0.000000, 0.000000, 0.000000}
\pgfsetstrokecolor{dialinecolor}
\node[anchor=west] at (5.746816\du,4.660000\du){};
\definecolor{dialinecolor}{rgb}{0.000000, 0.000000, 0.000000}
\pgfsetstrokecolor{dialinecolor}
\node[anchor=west] at (9.041816\du,5.10000\du){$R_2$};
\definecolor{dialinecolor}{rgb}{0.000000, 0.000000, 0.000000}
\pgfsetstrokecolor{dialinecolor}
\node[anchor=west] at (12.136816\du,5.10000\du){$R_3$};
\definecolor{dialinecolor}{rgb}{0.000000, 0.000000, 0.000000}
\pgfsetstrokecolor{dialinecolor}
\node[anchor=west] at (15.281816\du,5.10000\du){$R_d$};
\definecolor{dialinecolor}{rgb}{0.000000, 0.000000, 0.000000}
\pgfsetstrokecolor{dialinecolor}
\node[anchor=west] at (18.626816\du,5.10000\du){$R_{d+1}$};
\definecolor{dialinecolor}{rgb}{0.000000, 0.000000, 0.000000}
\pgfsetstrokecolor{dialinecolor}
\node[anchor=west] at (9.700000\du,5.150000\du){};
\definecolor{dialinecolor}{rgb}{0.000000, 0.000000, 0.000000}
\pgfsetstrokecolor{dialinecolor}
\node[anchor=west] at (7.405000\du,8.0000\du){$I_1$};
\definecolor{dialinecolor}{rgb}{0.000000, 0.000000, 0.000000}
\pgfsetstrokecolor{dialinecolor}
\node[anchor=west] at (10.690000\du,8.0000\du){$I_2$};
\definecolor{dialinecolor}{rgb}{0.000000, 0.000000, 0.000000}
\pgfsetstrokecolor{dialinecolor}
\node[anchor=west] at (16.805000\du,8.00000\du){$I_d$};
\end{tikzpicture}
\end{center}
\caption{Tensor train decomposition of a d-way tensor $\ten{A}$ into $d$ 3-way tensors $\ten{A}^{(1)}, \ten{A}^{(2)}\ldots,\ten{A}^{(d)}$.}
\label{fig:TTD}
\end{figure}
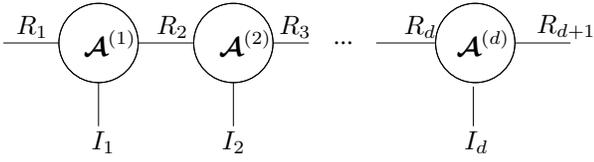

Here we briefly introduce the tensor train (TT) decomposition that will be utilized in the proposed K-STTM. A TT decomposition~\cite{oseledets2011tensor} represents a $d$-way tensor $\ten{A}$ as $d$ 3-way tensors $\ten{A}^{(1)}$, $\ten{A}^{(2)}$, \ldots , $\ten{A}^{(d)}$ such that a particular entry of $\ten{A}$ is written as the matrix product%
\begin{align}
\ten{A}(i_1,\ldots,i_d)&=\ten{A}^{(1)}(:,i_1,:)\cdots \ten{A}^{(d)}(:,i_d,:),
\label{eq:TT}
\end{align}
where $\ten{A}^{(k)}(:, i_k, :)$ is naturally a matrix since we fix the second index. Each tensor $\ten{A}^{(k)}$, $k=1,\ldots,d$, is called a \emph{TT-core} and has dimensions \mbox{$R_k \times I_k \times R_{k+1}$}. Storage of a tensor as a TT therefore reduces from $\prod_{i=1}^d\,R_i$ down to $\sum_{i=1}^d\,R_iI_iR_{i+1}$. In order for the left-hand-side of \eqref{eq:TT} to be a scalar we require that $R_1=R_{d+1}=1$. The remaining $R_k$ values are called the \emph{TT-ranks}. Figure~\ref{fig:TTD} demonstrates how TT-decomposition decomposes a $d$-way tensor $\ten{A}$, where the edges connecting the different circles indicate the matrix-matrix products of \eqref{eq:TT}. To simplify the statement, we define the notation $TT(\cdot)$, which means perform TT decomposition on a $d$-way tensor. For example, $TT(\ten{A})$ is the resulting TT after doing TT decomposition on the full tensor $\ten{A}$, namely, $\ten{A}^{(1)}$, $\ten{A}^{(2)}$, $\cdots$, $\ten{A}^{(d)}$ are derived.

\begin{figure}[t]
\begin{center}
\input{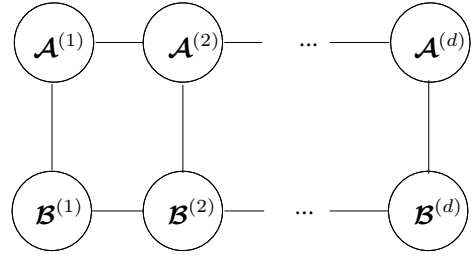}
\end{center}
\caption{The inner product between two $d$-way tensor trains.}
\label{fig:TT_inner_prod}
\end{figure}
\begin{definition}(TT inner product):
The inner product between two tensor trains $TT(\ten{A})$ and $TT(\ten{B})$ is denoted as $\langle TT(\ten{A}),TT(\ten{B}) \rangle$.
\end{definition}
The tensor network diagram of the inner product of two TTs is shown in Figure~\ref{fig:TT_inner_prod}. The lack of unconnected edges in Figure~\ref{fig:TT_inner_prod} implies that $\langle TT(\ten{A}),TT(\ten{B}) \rangle$ is a scalar.




\subsection{Support vector machines}
Since this work is based on traditional SVM, we therefore briefly review the main idea of an SVM. Assume we have a dataset~\mbox{$D$=\{$\mat{x_i}$, $y_i$\}$_{i=1}^M$} of $M$ labeled samples, where $\mat{x}_i \in \mathbb{R}^n$ are the vectorized data samples with labels $y_i\in \{-1,1\}$. The goal of an SVM is to find a discriminant hyperplane 
\begin{align}
  f(\mat{x})=\mat{w}^T\mat{x}+b
\label{eq:hyperplane}
\end{align}
that maximizes the margin between the two classes where $\mat{w}$ and $b$ are the weight vector and bias, respectively. However, an SVM is very sensitive to noise since it requires all the training samples to meet the hard margin constraint. In that case, the trained model tends to overfit. To solve this, slack variables $\xi_1,\ldots,\xi_M$ are introduced to allow some certain samples to be misclassified, thus enhancing the robustness of the trained model. We can express the learning problem as a quadratic optimization problem
\begin{align}
\label{eq:svmQP}
\nonumber \min_{\mat{w},b,\xi} 		&\quad \, \frac{1}{2} || \mat{w}||_F^2 +C\sum\limits_{i=1}^{M} \xi_i \\
\nonumber \textrm{subject to}  		&\quad \, y_i( \mat{w}^T \mat{x}_i+b )\ge 1-\xi_i,\\
									&\quad \, \xi_i\ge 0,\; i=1,\ldots ,M.
\end{align}
The parameter $C$ controls the trade-off between the size of the weight vector $\mat{w}$ and the size of the slack variables. It is more common to solve the dual problem of~\eqref{eq:svmQP}, especially when the feature size $n$ is larger than the sample size $M$. The dual problem format of~\eqref{eq:svmQP} is
\begin{align}
\nonumber \min\limits_{\alpha_1, \alpha_2, \cdots, \alpha_M} \quad \sum\limits_{i=1}^{M} \alpha_i - &\frac{1}{2} \sum\limits_{i,j=1}^{M} \alpha_i \alpha_j y_i y_j \langle \mat{x}_i, \mat{x}_j \rangle \\
\nonumber \textrm{subject to}\quad  &\sum\limits_{i=1}^{M}\alpha_i y_i = 0,\\  
0 \le \alpha_i\le C,\;& i=1,\ldots ,M,
\label{eq:dualSVM}
\end{align}
where $\langle \mat{x}_i, \mat{x}_j \rangle$ represents the inner product between vector $\mat{x}_i$ and $\mat{x}_j$ and $\alpha_i \, (i=1,\ldots ,M)$ are the Lagrange multipliers. 

To solve a nonlinear classification problem with SVM, researchers further introduced the kernel trick that projects the original vectorial data onto a much higher-dimensional feature space through a nonlinear mapping function $\phi$. In the feature space, the data generally become more (linearly) separable. By doing so, the optimization in~\eqref{eq:dualSVM} is transformed into
\begin{align}
\min\limits_{\alpha_1, \alpha_2, \cdots, \alpha_M} \quad \sum\limits_{i=1}^{M} \alpha_i - &\frac{1}{2} \sum\limits_{i,j=1}^{M} \alpha_i \alpha_j y_i y_j \langle \phi(\mat{x}_i), \phi(\mat{x}_j) \rangle 
\label{eq:dualnonlinearSVM}
\end{align}
with the same constraints as in~\eqref{eq:dualSVM}. It turns out that it is possible to make the computation easier by replacing the inner product term $\langle \phi(\mat{x}_i), \phi(\mat{x}_j) \rangle$ with a kernel function $\mat{k}(\mat{x}_i,\mat{x}_j)$. In that case, the inner product in the high-dimensional feature space can be computed without the need to explicitly compute the mappings $\phi(\mat{x}_i), \phi(\mat{x}_j)$. 

\section{KERNELIZED SUPPORT TENSOR TRAIN MACHINES}
\label{sec:ksttm}
In this section, we first demonstrate the tensor-based kernel learning problem and then introduce the proposed K-STTM.
\subsection{Problem statement}
Given $M$ tensorial training data and their labels, i.e., dataset $D=\{\ten{X}_i,y_i\}_{i=1}^M$, where \mbox{$\ten{X}_i\in \mathbb{R}^{I_1 \times I_2 \times \cdots \times I_d}$} and $y_i\in \{-1,1\}$, we want to find a hyperplane%
\begin{align}
  f(\ten{X})=\langle \ten{W},\ten{X} \rangle +b
\label{eq:Tensorhyperplane}
\end{align}
that separates the tensorial data into two classes. $\ten{W}$ is the hyperplane weight tensor with the same dimensions as $\ten{X}_i$ and $b$ is the bias. Similar to the primal problem in SVM, we can derive the corresponding primal optimization problem for ~\eqref{eq:Tensorhyperplane}
\begin{align}
\nonumber \min\limits_{\ten{W},b,\xi} \quad &\frac{1}{2}||\ten{W}||_F^2+C\sum\limits_{i=1}^{M} \xi_i\\
\nonumber \textrm{subject to}\quad  &y_i(\langle \ten{W}, \ten{X}_i \rangle +b)\ge1-\xi_i,\\
&\xi_i\ge 0,\; i=1,\ldots ,M.
\label{eq:sttmQPnew}
\end{align}
Following the scheme of the kernel trick for conventional SVMs, we introduce a nonlinear feature mapping function $\Phi$($\cdot$). Then, given a tensor $\ten{X} \in\mathbb{R}^{I_1\times I_2\times\cdots\times I_d}$, we assume it is mapped into the Hilbert space \ten{H} by
\begin{equation}
    \Phi:\ten{X} \rightarrow \Phi(\ten{X}) \in\mathbb{R}^{H_1\times H_2\times\cdots\times H_d}.
\end{equation}
We need to mention that the dimension of projected tensor $\Phi$($\ten{X}$) can be infinite depending on the feature mapping function $\Phi$($\cdot$). The resulting Hilbert space is then called the \emph{tensor feature space} and we can further develop the following model 
\begin{align}
\nonumber \min\limits_{\ten{W},b,\xi} \quad &\frac{1}{2}||\ten{W}||_F^2+C\sum\limits_{i=1}^{M} \xi_i\\
\nonumber \textrm{subject to}\quad  &y_i(\langle \ten{W}, \Phi (\ten{X}_i) \rangle +b)\ge1-\xi_i,\\
&\xi_i\ge 0,\; i=1,\ldots ,M,
\label{eq:primalSTM}
\end{align}
with parameter tensor $\ten{W} \in\mathbb{R}^{H_1\times H_2\times\cdots\times H_d}$. This model is naturally a linear classifier on the tensor feature space. However, when we map the classifier back to the original data space, it is a nonlinear classifier. To obtain the tensor-based kernel optimization model, we need to transfer model~\eqref{eq:primalSTM} into its dual, namely
\begin{align}
\nonumber \min\limits_{\alpha_1, \alpha_2, \cdots, \alpha_M} \quad \sum\limits_{i=1}^{M} \alpha_i - &\frac{1}{2} \sum\limits_{i,j=1}^{M} \alpha_i \alpha_j y_i y_j \langle \Phi(\ten{X}_i), \Phi(\ten{X}_j) \rangle \\
\nonumber \textrm{subject to}\quad  &\sum\limits_{i=1}^{M}\alpha_i y_i = 0,\\  
0 \le \alpha_i\le C,\;& i=1,\ldots ,M,
\label{eq:dualSTM}
\end{align}
where $\alpha_i$ are the Lagrange multipliers. The key task we need to solve is to define a tensorial kernel function $\ten{K}(\ten{X}_i,\ten{X}_j)$ that computes the inner product $\langle \Phi(\ten{X}_i), \Phi(\ten{X}_j) \rangle$ in the original data space instead of the feature space.

\subsection{Customized kernel mapping schemes for TT-based data}
\label{sec:reasonTT}
Although tensor is a natural structure for representing real-world data, there is no guarantee that such a representation works well for kernel learning. Instead of the full tensor, here we employ a TT for data representation due to the following reasons:
\begin{enumerate}
\item Real-life data often contain redundant information, which is not useful for kernel learning. The TT decomposition has proven to be efficient for removing the redundant information in the original data and provides a more compact data representation.
\item Compared to the Tucker decomposition whose storage scales exponentially with the core tensor, a TT is more scalable (parameter number grows linearly with the tensor order $d$), which reduces the computation during kernel learning.
\item Unlike the CP decomposition, determining the TT-rank is easily achieved through a series of singular value decompositions (TT-SVD~\cite{oseledets2011tensor} ). This naturally leads to a faster data transformation to the TT format. 
\item It is convenient to implement different operations on different tensor modes when data is in the TT format. Since a TT decomposition decomposes the original data into many TT cores, it is possible to apply different kernel functions on different TT cores for a better classification performance. Furthermore, we can emphasize the importance of different tensor modes by putting different weights on those TT cores during the kernel mapping. For example, a color image is a 3-way (pixel-pixel-color) tensor. The color mode can be treated differently with the two pixel modes since they contain different kinds of information, as will be exemplified later.

\end{enumerate}

In the following, we demonstrate the proposed TT-based feature mapping approach. Specifically, we map all fibers in each TT-core to the feature space, namely 
\begin{align}
   \nonumber  \Phi :\ten{X}^{(i)}(r_i,:,r_{i+1}) &\rightarrow  \Phi (\ten{X}^{(i)}(r_i,:,r_{i+1})) \in\mathbb{R}^{H_i} \\ 1 \le r_i\le R_i,&\quad  i=1,\ldots ,d,
\end{align}
where $\ten{X}^{(i)}$ and $R_i$ are the $i$-th TT-core and TT-rank of $TT(\ten{X})$, respectively. The fibers of each TT-core are naturally vectors, so the feature mapping works in the same way as for the conventional SVM. We then represent the resulting high-dimensional TT, which is in the tensor feature space, as $\Phi$($TT(\ten{X})$) $\in\mathbb{R}^{H_1\times H_2\times\cdots\times H_d}$. We stress that $\Phi$($TT(\ten{X})$) is still in a TT format with the same TT-ranks as $TT(\ten{X})$. In this sense, the TT format data structure is preserved after the feature mapping. 

After mapping the TT format data into the TT-based high-dimensional feature space, we then demonstrate the two proposed approaches for computing the inner product between two mapped TT format data using kernel function. 

\subsubsection{K-STTM-Prod}
The first method is called K-STTM-Prod since we implement consecutive multiplication operations on $d$ fiber inner products, which is consistent with the result of an inner product between two TTs. Assuming $\Phi$($TT(\ten{X})$) and $\Phi$($TT(\ten{Y})$) $\in\mathbb{R}^{H_1\times H_2\times\cdots\times H_d}$ with TT-ranks $R_i$ and $\hat{R}_i$, $i=1, 2, \ldots, d$, respectively, their inner product can be computed from
\begin{align}
\nonumber \langle \Phi( &TT(\ten{X})), \Phi(TT(\ten{Y})) \rangle = \sum\limits_{r_1 =1}^{R_1}\cdots \sum\limits_{r_d =1}^{R_d} \sum\limits_{\hat{r}_1 =1}^{\hat{R}_1}\cdots \sum\limits_{\hat{r}_d =1}^{\hat{R}_d} \\
&(\prod_{i=1}^d \langle \Phi (\ten{X}^{(i)}(r_i,:,r_{i+1})), \Phi (\ten{Y}^{(i)}(\hat{r}_i,:,\hat{r}_{i+1})) \rangle).
\label{eq:innerprodK_STTM}
\end{align}
We remark that~\eqref{eq:innerprodK_STTM} derives the exact same result as Figure~\ref{fig:TT_inner_prod} (assuming $\ten{X}=\ten{A}$ and $\ten{Y}=\ten{B}$) when an identity feature mapping function $\Phi$($\cdot$) is used, namely $\Phi$($TT(\ten{X})$)=$TT(\ten{X})$. What is more, since each fiber of a mapped TT-core is naturally a vector, we have 
\begin{align}
\nonumber \langle \Phi (\ten{X}^{(i)}(r_i,:,r_{i+1})), \Phi (\ten{Y}^{(i)}(\hat{r}_i,:,\hat{r}_{i+1})) \rangle = \\
\ten{K} (\ten{X}^{(i)}(r_i,:,r_{i+1}),  \ten{Y}^{(i)}(\hat{r}_i,:,\hat{r}_{i+1})),
\label{eq:kerneltrickK_STTM}
\end{align}
where $\ten{K}$($\cdot$) can be any kernel function used for a standard SVM, such as a Gaussian RBF kernel, polynomial kernel, linear kernel etc. Combining~\eqref{eq:innerprodK_STTM} and~\eqref{eq:kerneltrickK_STTM}, we obtain the corresponding TT-based kernel function 
\begin{align}
\nonumber \ten{K}(TT(\ten{X}),TT(\ten{Y})) = \sum\limits_{r_1 =1}^{R_1}\cdots \sum\limits_{r_d =1}^{R_d} \sum\limits_{\hat{r}_1 =1}^{\hat{R}_1}\cdots \sum\limits_{\hat{r}_d =1}^{\hat{R}_d}\\
(\prod_{i=1}^d \ten{K} (\ten{X}^{(i)}(r_i,:,r_{i+1}),  \ten{Y}^{(i)}(\hat{r}_i,:,\hat{r}_{i+1}))).
\label{eq:kernelfuncK_STTM_prod}
\end{align}
As mentioned before, different kernel functions can be applied on different tensor modes $i=1,2,\ldots,d$. Therefore, the second line in~\eqref{eq:kernelfuncK_STTM_prod} can be generalized to 
\begin{equation}
\nonumber (\prod_{i=1}^d \ten{K}_i (\ten{X}^{(i)}(r_i,:,r_{i+1}),  \ten{Y}^{(i)}(\hat{r}_i,:,\hat{r}_{i+1}))).
\end{equation}
One possible application is in color image classification, where one could apply Gaussian RBF kernels $\ten{K}_1$ and $\ten{K}_2$ on its first two spatial modes, while choosing a linear or polynomial kernel $\ten{K}_3$ for the color mode. This will be investigated in the experiments.   

\subsubsection{K-STTM-Sum}
The second method we propose to construct a TT kernel function is called K-STTM-Sum. Instead of implementing consecutive multiplication operations on $d$ fiber inner products like in K-STTM-Prod, K-STTM-Sum performs consecutive addition operations on them. This idea is inspired by~\cite{tensormultiview} which argues that the product of inner products can lead to the loss/misinterpretation of information. Take the linear kernel as an example, the inner product between two fibers of the same mode could be negative, which indicates a low similarity between those two fibers. However, by implementing consecutive multiplication operations on $d$ fiber inner products, highly negative values could result in a large positive value. In that case, the overall similarity is high which is clearly unwanted. This situation also appears when employing Gaussian RBF kernels. A nearly zero value would be assigned to two non-similar fibers, which could influence the final result significantly. To this end, we propose the K-STTM-Sum. Similar to K-STTM-Prod, we can obtain the corresponding kernel function as%
\begin{align}
\nonumber \ten{K}(TT(\ten{X}),TT(\ten{Y})) = \sum\limits_{r_1 =1}^{R_1}\cdots \sum\limits_{r_d =1}^{R_d} \sum\limits_{\hat{r}_1 =1}^{\hat{R}_1}\cdots \sum\limits_{\hat{r}_d =1}^{\hat{R}_d}\\
(\sum_{i=1}^d \ten{K}_i(\ten{X}^{(i)}(r_i,:,r_{i+1}),  \ten{Y}^{(i)}(\hat{r}_i,:,\hat{r}_{i+1}))).
\label{eq:kernelfuncK_STTM_sum}
\end{align}

\subsection{Kernel optimization problem}
After defining the TT-based kernel function, we can then replace the term $\langle \Phi(\ten{X}_i), \Phi(\ten{X}_j) \rangle$ in~\eqref{eq:dualSTM} with~\eqref{eq:kernelfuncK_STTM_prod} or~\eqref{eq:kernelfuncK_STTM_sum}, and derive our final kernel optimization problem based on the TT structure, namely,%
\begin{align}
\nonumber \min\limits_{\alpha_1, \alpha_2, \cdots, \alpha_M}  \sum\limits_{i=1}^{M} \alpha_i - &\frac{1}{2} \sum\limits_{i,j=1}^{M} \alpha_i \alpha_j y_i y_j \ten{K}(TT(\ten{X}_i),TT(\ten{X}_j)) \\
\nonumber \textrm{subject to}\quad  &\sum\limits_{i=1}^{M}\alpha_i y_i = 0,\\  
0 \le \alpha_i\le C,\;& i=1,\ldots ,M.
\label{eq:dualKSTTM}
\end{align}

After solving~\eqref{eq:dualKSTTM}, we can get the unknown model parameters $\alpha_1, \alpha_2, \ldots, \alpha_M$ and the resulting decision function is then represented as
\begin{equation}
    f(\ten{X})=\textrm{sign}(\sum\limits_{i=1}^{M}\alpha_i y_i \ten{K}(TT(\ten{X}_i),TT(\ten{X}))+b).
\end{equation}
Here we take the Gaussian RBF kernel as an example and summarize the training algorithm of the K-STTM-Prod/Sum as pseudo-code in Algorithm~\ref{alg:KSTTM}. An alternative for doing a grid search to find optimal hyperparameters would be cross-validation. Generalizing the binary classification to multi-classification can be easily achieved by utilizing a one-vs-one or one-vs-all strategy, namely, we can build several binary classifiers to do multi-class classification.

\begin{algorithm}[t]
	\renewcommand{\algorithmicrequire}{\textbf{Input:}}
	\renewcommand{\algorithmicensure}{\textbf{Output:}}
	\caption{K-STTM-Prod/Sum Algorithm}
	\label{alg:KSTTM}
	\begin{algorithmic}[1]
		\REQUIRE Training dataset $\{\ten{X}_i\in\mathbb{R}^{I_1\times \cdots\times I_d}$, $y_i\in \{-1, 1 \}$\}$_{i=1}^M$; Validation dataset $\{\ten{X}_j\in\mathbb{R}^{I_1\times \cdots\times I_d}$, $y_j\in \{-1, 1 \}$\}$_{j=1}^{N}$; The pre-set TT-ranks $R_1, R_2, \ldots, R_{d+1}$; The range of the performance trade-off parameter $C$ and kernel width parameter $\sigma$, namely $[C_{min}, C_{max}]$, and $[\sigma _{min}, \sigma _{max}]$.     
		\ENSURE The Lagrange multipliers $\alpha_1, \alpha_2, \dots, \alpha_M$; The bias $b$.
        \vspace{1ex}
		\STATE Compute the TT approximation of training samples $\{\ten{X}_i\}_{i=1}^M$ and validation set $\{\ten{X}_j\}_{j=1}^N$ with the given TT-ranks using TT-SVD.
		\FOR{$C$ from $C_{min}$ to $C_{max}$}
		\FOR{$\sigma$ from $\sigma_{min}$ to $\sigma_{max}$}
        \STATE Apply Gaussian RBF kernel on all the tensor modes, and construct the kernel matrix according to~\eqref{eq:kernelfuncK_STTM_prod} or~\eqref{eq:kernelfuncK_STTM_sum}, which are corresponding to K-STTM-Prod and K-STTM-Sum, respectively.      
        \STATE Solve~\eqref{eq:dualKSTTM} using the resulting kernel matrix.
        \STATE Compute the classification accuracy on validation set.
        \ENDFOR	
        \ENDFOR
        \STATE Find the best $C$ and $\sigma$ according to the classification accuracy on validation set.
        \STATE Train the K-STTM with the best $C$ and $\sigma$ by implementing step $4$ and $5$. Thus the the Lagrange multipliers $\alpha_1, \alpha_2, \dots, \alpha_M$ and the bias $b$ are obtained.
	\end{algorithmic}%
\end{algorithm}%



\subsection{Kernel validity}
\label{sec:kernel_valid}
A key property of kernel function in standard SVM is that the resulting kernel matrix is positive semi-definite, which guarantees the mapped high-dimensional feature space is truly an inner product space. Therefore, we provide Theorem~\ref{theo_validity} to show the validity of K-STTM-Prod and K-STTM-Sum.

\begin{theorem}
\label{theo_validity}
Kernel functions K-STTM-Prod and K-STTM-Sum produce positive semi-definite kernel matrices.\\
\end{theorem}
We provide the proof here.
\subsubsection{Validity of K-STTM-Prod}
We first demonstrate the kernel function validity of K-STTM-Prod. The goal is to show that the final kernel matrix constructed by~\eqref{eq:kernelfuncK_STTM_prod} is positive semi-definite. In the actual implementation, it is extremely inefficient to use TT decomposition to decompose each tensorial sample one by one. The way we did it is by first stacking all the $d$-way samples and then compute a TT decomposition on the resulting $(d+1)$-way tensor directly. By doing so, all TT-based training samples have the same TT-ranks. Also in the case where we compute the TT decomposition separately for each sample, we can still set the TT-ranks of all samples to be identical. That means $R_i$ is equal to $\hat{R}_i$, $i=1,2,\ldots,d$ for all the TT-based training samples. We can then compute the final kernel matrix by doing $R_1^2\times \ldots \times R_d^2$ matrix summations, while each matrix in the summation procedure is computed by $d$ times matrix Hadamard product. The matrix factors in the $d$ times Hadamard product are valid kernel matrices since they are computed using the standard kernel function.

Through the above analysis, the goal now transforms into proving that the summation or Hadamard product between two positive semi-definite matrices $\mat{A}$ and $\mat{B}$ $\in\mathbb{R}^{n\times n} $ still results in a positive semi-definite matrix. 

For the summation case, we have 
\begin{equation}
\label{summation1}
\left\{
\begin{aligned}
\mat{u}^T\mat{A}\mat{u} \geq 0, \\
\mat{u}^T\mat{B}\mat{u} \geq 0.
\end{aligned}
\right.
\end{equation}
for every non-zero column vector $\mat{u}$ $\in\mathbb{R}^{n}$. Obviously we can conclude that 
\begin{equation}
\label{summation1}
\mat{u}^T\mat{(A+B)}\mat{u} \geq 0, 
\end{equation}
namely $\mat{A}+\mat{B}$ is still positive semi-definite. 

For the Hadamard product case, we refer to the the Schur product theorem~\cite{schur1911bemerkungen} and we can easily obtain 
\begin{equation}
\label{summation1}
\mat{u}^T\mat{(A\odot  B)}\mat{u} \geq 0, 
\end{equation}
for every non-zero column vector $\mat{u}$ $\in\mathbb{R}^{n}$, where $\odot$ is the Hadamard product. Thus $\mat{A\odot  B}$ is still positive semi-definite.

Through the above analysis, we can conclude that by constraining the TT-based training samples to have identical TT-ranks, we can get a valid kernel matrix using K-STTM-Prod.

\subsubsection{Validity of K-STTM-Sum}

The proof for the validity of K-STTM-Sum is similar as it for K-STTM-Prod. The difference between kernel functions~\eqref{eq:kernelfuncK_STTM_prod} and~\eqref{eq:kernelfuncK_STTM_sum} is that~~\eqref{eq:kernelfuncK_STTM_sum} only replaces the product with a summation. In that case, for K-STTM-Sum, the final kernel matrix is produced by the summation of a set of valid positive semi-definite matrices. Namely, we can still get a valid kernel matrix using K-STTM-Sum.

\subsection{Convergence and complexity}
\label{sec:analysis}


Here we demonstrate the convergence analysis of our proposed methods and compare the storage and computation complexity with the standard SVM.

For the convergence analysis, it is same as it in standard SVM problem. 
We already show the kernel validity of~\eqref{eq:kernelfuncK_STTM_prod} and~\eqref{eq:kernelfuncK_STTM_sum} in Theorem~\ref{theo_validity}. With a valid kernel matrix, we can solve a quadratic programming problem to get the Lagrange multipliers $\alpha_i$ and bias $b$, which is same as the procedure in standard SVM. Consequently, the convergence analysis is exactly same as it in standard SVM. 

For the storage complexity analysis, the original tensorial sample storage is $O(MI^d)$, where $I$ is the maximum value of $I_i,~i=1,2,\ldots, d$. After representing the original tensorial data as TTs, the data storage becomes to $O(MdIR^2)$, where $R$ is the maximum TT-rank. This shows a great reduction especially when the data order $d$ is large.

For the computation complexity, the overall result of K-STTM-Prod is the same as the result of K-STTM-Sum if we neglect those low-order polynomial terms. This can be observed from~\eqref{eq:kernelfuncK_STTM_prod} and~\eqref{eq:kernelfuncK_STTM_sum}. The main computation costs are similar in those two equations. Therefore we just analyze the K-STTM-Prod method. The computational complexity of constructing the kernel matrix in standard SVM is $O(M^2I^d)$, where $n$ is the maximum dimension of $I_i,~i =1,2,...d$. When applying the accelerating implementation of K-STTM-Prod, its kernel matrix computation complexity is $O(dI^2R^4+M^2I_dR_d^2)$, where $I$ and $R$ are the maximum values of $I_i$ and $R_i$, $i =1,2,...d-1$, respectively. Real-world data is commonly low-rank, so the TT-ranks $R_i$ are generally small. Moreover, their dimensions $I_i$ are very high. That indicates our proposed method is more efficient than its vector counterpart since the computation complexity is reduced from exponential to polynomial.

\section{EXPERIMENTS}
\label{sec:experiments}
 We evaluate the effectiveness of the two proposed schemes, K-STTM-Prod and K-STTM-Sum, on real-world tensorial datasets and contrast our methods with the following seven methods as a baseline.
\begin{itemize}
\item {\bf{SVM}}: SVM~\cite{boser1992training} is one of the most widely used vector-based method for classification. What is more, the proposed K-STTM is a tensorial extension of SVM, so SVM is selected as a baseline. We employ the widely used convex optimization solver CVX\footnote{http://cvxr.com/cvx/} to solve the quadratic programming problem.
\item {\bf{STM}}: STM~\cite{STM} is the first method which extends SVM to the tensorial format, which employs alternating optimization scheme to update the weight tensors and outperforms kernel SVM in some tasks.
\item {\bf{STuM}}: It is a kind of support tensor machine which is based on the Tucker decomposition~\cite{kotsia2011support}. The training procedure is similar as the one in STM. 
\item {\bf{STTM}}: STTM~\cite{linearSTTM} assumes the weight tensor is a scalable tensor train, which enables STTM to deal with high-dimensional data classification. STM, STuM, and STTM are all tensor-based linear classifiers. In very small sample size problem, sometimes linear classifier are observed to achieve a better classification accuracy than nonlinear classifier~\cite{he2014dusk} since a linear classifier is commonly less complex and more stable and can be better trained than nonlinear classifiers. 
\item {\bf{DuSK}}: DuSK is a kernelized support tensor machine using the CP decomposition~\cite{he2014dusk}. Through introducing the kernel trick, it can deal with nonlinear classification tasks. 
\item {\bf{3D CNN}}: CNN is one of the most powerful structure for image classification. The 3D CNN we employ here is an extension of the 2D version in~\cite{gupta2013natural}. We replace the 2D convolutional kernels with 3D ones and keep other settings the same. Though 3D CNN is a relatively simple CNN model, it has an advantage in dealing with small sample size problems since it can be trained better than the complicated CNN model.   
\item {\bf{TT Classifier}}: As an updated tensor classification method, TT classifier~\cite{chen2017parallelized} trains a TT as a polynomial classifier and achieves good results on tensorial image classification tasks.
\end{itemize}

    For simplicity, all of the kernel based methods, i.e., SVM, DuSK, and K-STTM, employ the Gaussian RBF kernel. 
    The optimal parameters, namely the performance trade-off parameter $C$, RBF kernel parameter $\sigma$, hidden layer size in 3D CNN,
    plus the corresponding tensor rank in STuM, STTM, DuSK, TT classifier and K-STTM,  are determined through a grid search. The detail of the hyperparameter search schemes of all methods in all experiments are demonstrated in Appendix.  

\begin{table*}[tbh]
\newcommand{\tabincell}[2]{\begin{tabular}{@{}#1@{}}#2\end{tabular}}
\centering
\caption{\label{mnistcomparison} Classification accuracy of different methods for different MNIST digit pairs.}
\vspace{1ex}
\begin{tabular}{@{}lrrrrrrrrr@{}}
Digit pair &SVM & STM & STuM &STTM & DuSK & 3D CNN & TT classiifer &K-STTM-Prod & K-STTM-Sum \\
\midrule
\tabincell{c}{\{`1',`2'\}}  & $98.15\%$& $94.09\%$& $97.96\%$& $97.69\%$&$89.16\%$ & $98.20\%$& $73.93\%$& $99.22\%$ & $\textbf{99.27\%}$  \\
\tabincell{c}{\{`1',`7'\}}  & $97.73\%$ &$96.62\%$ &$97.87\%$ &$97.83\%$ &$80.95\%$ & $98.19\%$& $96.99\%$& $\textbf{98.34\%}$&$98.20\%$  \\
\tabincell{c}{\{`1',`8'\}}  & $96.49\%$& $93.78\%$& $95.92\%$& $96.30\%$&$87.63\%$ & $97.24\%$& $84.48\%$& $97.97\%$& $\textbf{98.06\%}$   \\
\tabincell{c}{\{`2',`4'\}}  & $98.36\%$&$96.32\%$ &$97.46\%$ &$97.52\%$ &$77.26\%$ & $96.27\%$& $93.45\%$& $\textbf{99.01\%}$& $98.61\%$   \\

\tabincell{c}{\{`2',`7'\}}  & $96.41\%$&$94.46\%$ &$95.58\%$ &$95.58\%$ &$81.26\%$ & $94.22\%$& $94.22\%$& $\textbf{97.09\%}$&$96.95\%$ \\
\tabincell{c}{\{`4',`6'\}}  & $97.16\%$ &$97.57\%$ &$97.47\%$ &$97.42\%$ &$78.66\%$ & $96.18\%$& $93.71\%$& $\textbf{98.30\%}$& $97.74\%$ \\ 
\tabincell{c}{\{`4',`9'\}}  & $89.50\%$&$86.53\%$ &$90.65\%$ &$90.86\%$ & $68.46\%$& $91.91\%$& $59.12\%$& $\textbf{93.27\%}$& $91.77\%$    \\ 
\tabincell{c}{\{`5',`6'\}}  & $96.00\%$& $95.29\%$& $95.24\%$& $94.92\%$& $75.78\%$& $92.75\%$& $88.16\%$& $96.49\%$& $\textbf{96.76\%}$  \\
\tabincell{c}{\{`5',`8'\}}  & $86.92\%$&$78.18\%$ & $91.47\%$& $88.10\%$& $70.69\%$& $90.46\%$& $63.56\%$& $\textbf{94.32\%}$& $91.59\%$ \\
\tabincell{c}{\{`7',`8'\}}  & $94.46\%$&$92.30\%$ &$95.85\%$ & $95.40\%$&$75.97\%$ & $94.30\%$& $94.46\%$& $\textbf{96.76\%}$& $96.16\%$   \\

\end{tabular}
\label{exp1_tbl2}
\end{table*}

\subsection{MNIST}
\label{exp_1}
    First, our proposed methods are compared with the above methods on the well-known MNIST dataset~\cite{lecun-mnisthandwrittendigit-2010}, which has a training set of $60$k samples and a testing set of $10$k samples. Each sample is a $28\times 28$ grayscale image of a handwritten digit \{$0,\ldots,9$\}. Although there are total $60$k training samples, we care more about the small sample size problem. Thus for each class, we randomly choose $50$ samples for model training and another $50$ for validation. All test samples in each class are used for checking the classification performance of each trained model. Since an SVM is naturally a binary classifier, we randomly choose $10$ digit pairs out of $45$ to check the classification accuracy.


Table~\ref{exp1_tbl2} shows the classification results on different digit pairs. DuSK achieves the lowest accuracy among all the methods , which may be caused by the  CP-rank searching: Finding a good CP-rank is an NP-complete problem, so DuSK may perform poorly if it fails to do so. 
Due to the naturally linearity of STM, STuM and STTM, they also can not achieve a good classification accuracy on real-world data. TT classifier even achieves a very poor classification performance on some digit pairs since it is naturally a polynomial classifier, whose classification power is limited. We notice that the two proposed approaches K-STTM-Prod and K-STTM-Sum only achieve a slightly better accuracy than SVM and 3D CNN on some digit pairs. The main reason that SVM and 3D CNN perform very well also is that MNIST is a relatively small dataset. The data dimension is $784$ only, thus conventional SVM and 3D CNN do not encounter the curse of dimensionality and no overfitting occurs. The advantage of tensorial methods are expected to be more apparent when the problem is truly high-dimensional. We therefore consider in the second experiment fMRI image data, whose dimensions are higher than $32$k. 

\begin{figure}[tb]
    \centering
    \subfigure{    \includegraphics[width=3.7in]{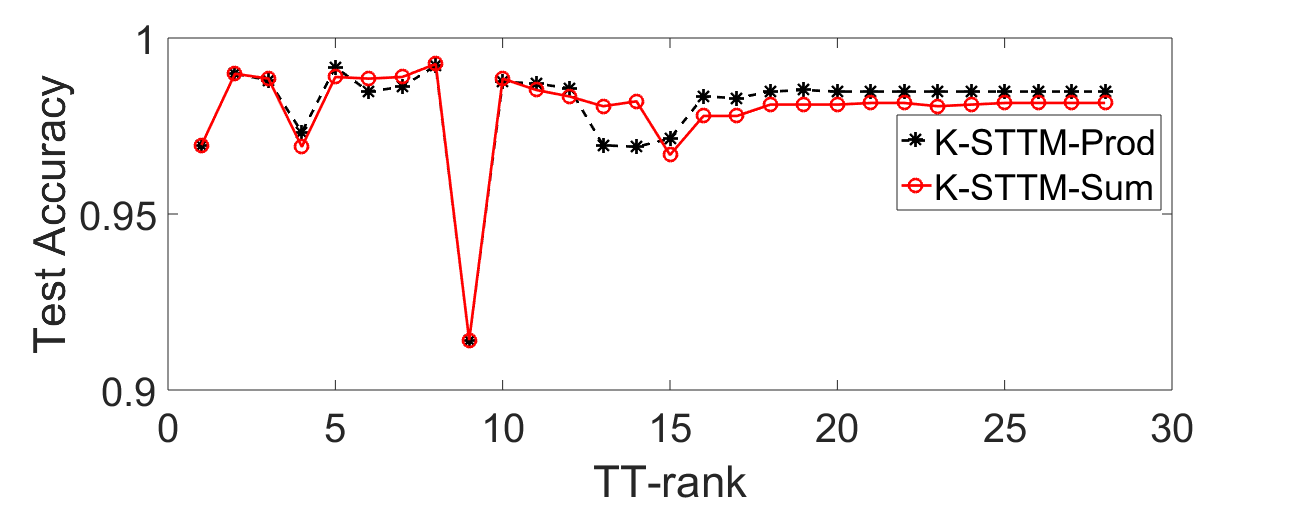}
    }
    \quad
    \subfigure{       \includegraphics[width=3.7in]{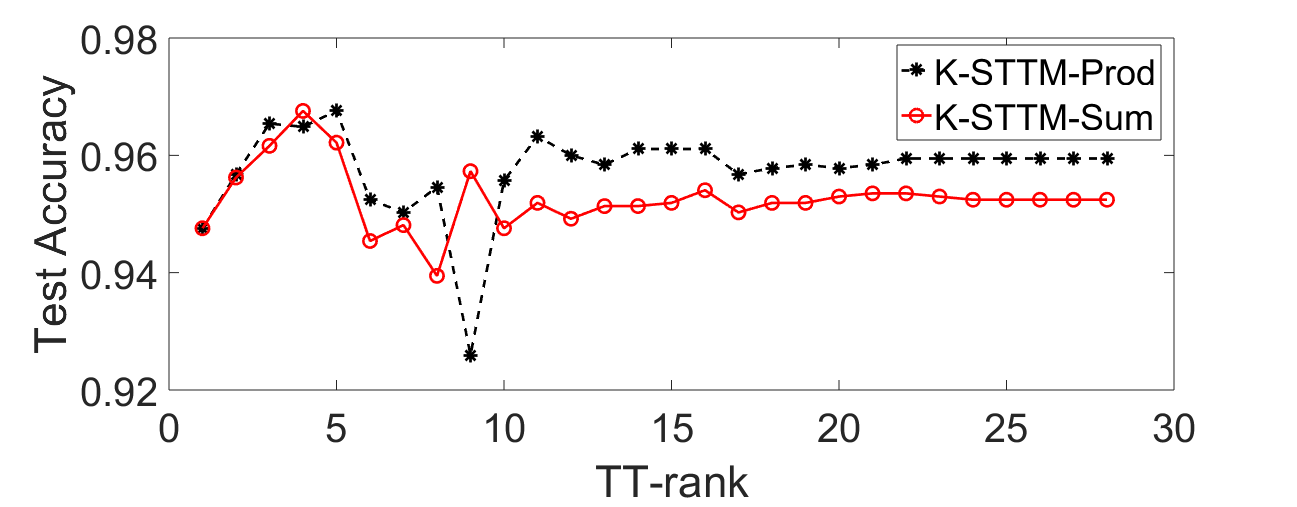}
    }
    \caption{Classification accuracy of K-STTM-Prod and K-STTM-Sum with different TT-rank on two randomly selected digit pairs. Top figure: digit pair `1',`2'; bottom figure: digit pair `5',`6'.}
    \label{fig:diffRank}
\end{figure}

We also investigate the influence of the TT-rank on the classification accuracy. Figure~\ref{fig:diffRank} shows how the classification accuracy of K-STTM-Prod and K-STTM-Sum changes along with increasing TT-rank on two randomly selected digit pairs. We can observe that K-STTM with a small TT-rank can achieve a similar classification performance when it with high TT-rank, and the highest accuracies are all achieved when TT-rank is around 5, which means we can select a relatively small TT-rank $R$ to reduce the cost of kernel computation, while at the same time keep the classification performance. This validates the computation complexity analysis in Section~\ref{sec:analysis} since the $R$ and $R_d$ in 
$O(dI^2R^4+M^2I_dR_d^2)$ are often small.

\begin{table*}[tbh]
\newcommand{\tabincell}[2]{\begin{tabular}{@{}#1@{}}#2\end{tabular}}
\centering
\caption{\label{starpluscomparison} Classification accuracy of different methods for different subjects in StarPlus fMRI datasets.}
\vspace{1ex}
\begin{tabular}{@{}lrrrrrrrrr@{}}
Subject &SVM & STM & STuM &STTM & DuSK & 3D CNN & TT classifier & K-STTM-Prod & K-STTM-Sum \\
\midrule
\tabincell{c}{04799}  & $50.00\%^*$& $36.67\%$& $35.83\%$& $39.61\%$&$47.50\%$ & $51.67\%$& $57.50\%$  & $68.33\%$ & $\textbf{66.67\%}$  \\
\tabincell{c}{04820}  & $50.00\%^*$ &$43.33\%$ &$35.00\%$ &$45.83\%$ &$46.67\%$ & $44.16\%$& $54.17\%$ & $\textbf{70.00\%}$&$62.50\%$  \\
\tabincell{c}{04847}  & $50.00\%^*$& $38.33\%$& $17.50\%$& $47.50\%$&$53.33\%$ & $55.00\%$& $61.67\%$  &$65.00\%$& $\textbf{65.00\%}$   \\
\tabincell{c}{05675}  & $50.00\%^*$&$37.50\%$ &$30.83\%$ &$35.00\%$ &$55.00\%$ & $47.50\%$& $55.00\%$ & $\textbf{60.00\%}$& $60.00\%$   \\

\tabincell{c}{05680}  & $50.00\%^*$&$38.33\%$ &$39.17\%$ &$40.00\%$ &$64.17\%$ & $68.33\%$& $60.83\%$ & $\textbf{73.33\%}$&$75.00\%$ \\
\tabincell{c}{05710}  & $50.00\%^*$ &$40.00\%$ &$30.00\%$ &$43.33\%$ &$54.16\%$ & $47.50\%$& $53.33\%$  & $\textbf{59.17\%}$& $58.33\%$ \\ 

\end{tabular}
\label{exp2_tbl1}
\begin{tablenotes}
        \item[*] $~~~~~~~~~~~^*$ SVM classifies all test samples into one class since no good hyperparameter setting can be found by grid search.
        \end{tablenotes}
\end{table*}

\subsection{fMRI datasets}
\label{exp_2}
\begin{figure}[t]
\centering
\includegraphics[width=3.3in]{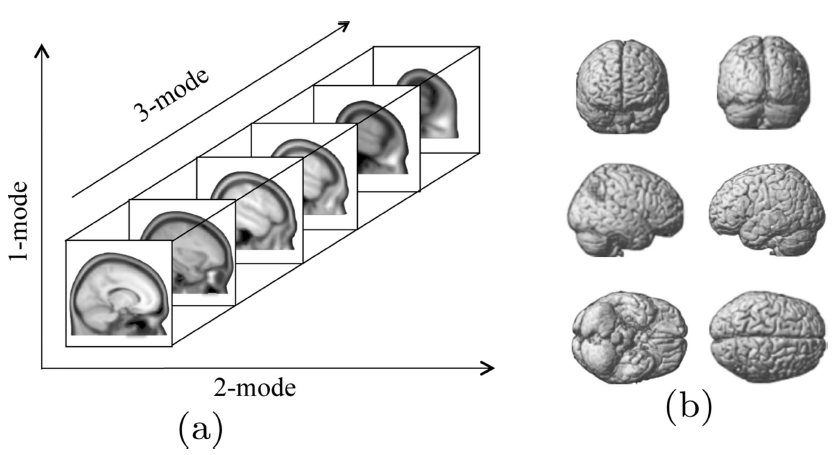}
\caption{fMRI images from~\cite{he2014dusk}. (a) An illustration of a 3-way tensor (fMRI image), (b) Visualization of an fMRI image.}
\label{fig:fMRI}
\end{figure}

As we mentioned in experiment~\ref{exp_1}, tensorial method shows more apparent advantages on high-dimensional dataset. Thus we consider two high-dimensional fMRI datasets, namely the StarPlus fMRI dataset\footnote{http://www.cs.cmu.edu/afs/cs.cmu.edu/project/theo-81/www/} and the CMU Science 2008 fMRI dataset (CMU2008)~\cite{mitchell2008predicting} to evaluate the classification performance of different models. An fMRI image is essentially a 3-way tensor. Figure~\ref{fig:fMRI} from~\cite{he2014dusk} illustrates the tensorial structure of the fMRI image. 
\subsubsection{StarPlus fMRI dataset}
\label{exp_2_1}
The fMRI images in StarPlus dataset are with dimensions $64\times 64 \times 8$ that contains 25 to 30 anatomically defined regions (called``Regions of Interest'', or ROIs). To achieve a better classification accuracy, we only consider the following ROIs: {`CALC' `LIPL' `LT' `LTRIA' `LOPER' `LIPS' `LDLPFC'}. After extracting those ROIs, we further normalize the data of each subject.  
StarPlus fMRI dataset contains the brain images of $6$ human subjects. The data of each human subject is partitioned into trials, and each subject has $40$ effective trials.
Here we only use the first 4 seconds of each trial since the subject was shown one kind of stimulus (sentence or picture) during the whole period. The fMRI images were collected every 500 msec, thus we can utilize $8$ fMRI images in each trial. Overall, we have $320$ fMRI images: one half of them were collected when the subject was shown a picture, the other half were collected when the subject was shown a sentence, while we randomly select 140 images for training, 60 for validation and the left for testing. 


The classification results are listed in Table~\ref{exp2_tbl1}. Due to the very high-dimensional and sparse data, SVM fails to find a good hyperparameter setting thus can not do classification. Since fMRI data are very complicated, those linear classifiers, namely STM, STuM and STTM, can not achieve an acceptable performance, and the classification accuracies of them are all lower than $50\%$. The classification result of TT classifier is poor on several subjects. DuSK also performs poor on subjects `04799' and `04820'. Due to the small number of training samples and high-dimensional data size, the 3D CNN overfits and can not be well trained, while our proposed two methods still achieve the highest classification accuracy on all human subjects.

\begin{table*}[t]
\newcommand{\tabincell}[2]{\begin{tabular}{@{}#1@{}}#2\end{tabular}}
\centering
\caption{\label{cmu2008comparison} Classification accuracy of different methods for different subjects in CMU2008 fMRI datasets.}
\vspace{1ex}
\begin{tabular}{@{}lrrrrrrrrr@{}}
Subject &SVM & STM & STuM &STTM & DuSK & 3D CNN & TT classifier & K-STTM-Prod & K-STTM-Sum \\
\midrule
\tabincell{c}{~~~~\#1}  & $68.00\%$ &$66.00\%$ &$68.00\%$ &$66.00\%$ &$48.00\%$ & $52.00\%$& $28.00\%$ & $\textbf{70.00\%}$&$\textbf{70.00\%}$  \\ 
\tabincell{c}{~~~~\#2}  & $52.00\%$& $50.00\%$& $58.00\%$& $68.00\%$&$54.00\%$ & $58.00\%$&  $44.00\%$ &$74.00\%$& $\textbf{84.00\%}$   \\

\tabincell{c}{~~~~\#3}  & $50.00\%$&$60.00\%$ &$58.00\%$ &$64.00\%$ &$58.00\%$ & $56.00\%$& $60.00\%$ & $66.00\%$&$\textbf{70.00\%}$ \\
\tabincell{c}{~~~~\#4}  & $50.00\%$ &$60.00\%$ &$58.00\%$ &$56.00\%$ &$52.00\%$ & $58.00\%$& $56.00\%$  & $\textbf{76.00\%}$& $72.00\%$ \\ 
\tabincell{c}{~~~~\#5}  & $56.00\%$&$58.00\%$ &$64.00\%$ &$66.00\%$ & $44.00\%$& $60.00\%$& $56.00\%$ & $\textbf{72.00\%}$& $\textbf{72.00\%}$    \\ 
\tabincell{c}{~~~~\#6}  & $44.00\%$& $60.00\%$& $46.00\%$& $46.00\%$& $54.00\%$& $56.00\%$& $48.00\%$  & $\textbf{70.00\%}$& $\textbf{70.00\%}$  \\
\tabincell{c}{~~~~\#7}  & $50.00\%$&$52.00\%$ & $48.00\%$& $52.00\%$& $52.00\%$& $60.00\%$& $60.00\%$  & $68.00\%$& $\textbf{72.00\%}$ \\ 

\end{tabular}
\label{exp2_tbl2}
\end{table*}

\begin{table*}[t]
\newcommand{\tabincell}[2]{\begin{tabular}{@{}#1@{}}#2\end{tabular}}
{ 
\centering
\caption{\label{cifarcomparison}Classification accuracy of SVM, STM, STuM, STTM, DuSK, 3D CNN, TT classiifer and K-STTM-Prod with different kernel functions (the last 3 columns) for different CIFAR-10 class pairs.}
\begin{tabular}{@{}lrrrrrrrrrr@{}}
\tabincell{c}{Class pair}&SVM& STM & STuM&STTM & DuSK & 3D CNN &\tabincell{c}{TT\\classifier}  &\tabincell{c}{RBF-RBF\\-RBF} & \tabincell{c}{RBF-RBF\\-Poly.}& \tabincell{c}{RBF-RBF\\-Linear}\\

\midrule
\tabincell{c}{`automobile',`cat'}& $76.05\%$ & $49.85\%$ & $68.15\%$ & $68.90\%$  & $61.90\%$ & $76.75\%$& $49.60\% $ & $77.15\% $& $77.10\%$ & $\textbf{77.75\%}$  \\
\tabincell{c}{`automobile',`frog'}& $83.85\%$& $48.15\%$ & $69.90\%$ & $70.95\%$  & $68.85\%$& $79.45\%$ & $49.40\%$  & $85.15\%$& $85.20\%$ & $\textbf{85.30\%}$   \\
\tabincell{c}{`automobile',`ship'}& $77.20\%$& $55.30\%$ & $71.40\%$ & $69.40\%$   & $64.70\%$& $74.40\%$ & $52.05\%$  & $78.90\%$& $79.25\%$ & $\textbf{80.40\%}$ \\
\tabincell{c}{`bird',`deer'} & $58.20\%$ & $49.90\%$ & $53.65\%$ & $58.70\%$   & $56.85\%$& $58.90\%$ & $47.80\%$ & $62.95\%$& $63.05\%$ & $\textbf{63.55\%}$  \\
\tabincell{c}{`cat',`ship'} & $80.35\%$ & $56.80\%$ & $80.25\%$ & $76.45\%$    & $72.00\%$ & $80.00\%$ & $49.10\%$  & $\textbf{81.25\%}$& $80.10\%$ & $81.05\%$ \\
\tabincell{c}{`deer',`frog'}& $61.80\%$ & $48.25\%$ & $60.65\%$ & $60.40\%$ & $57.35\%$& $67.85\%$& $49.40\%$   & $68.40\%$& $68.65\%$ & $\textbf{69.00\%}$ \\
\tabincell{c}{`deer',`horse'}& $72.50\%$& $50.60\%$ & $59.45\%$ & $62.15\%$   & $63.30\%$& $66.65\%$& $50.80\%$   & $72.80\%$& $72.90\%$ & $\textbf{73.00\%}$  \\
\tabincell{c}{`deer',`ship'}& $82.35\%$ & $57.80\%$ & $83.50\%$ & $81.45\%$ & $75.05\%$ & $82.20\%$&  $50.30\%$  & $83.65\%$& $\textbf{83.90\%}$ & $83.40\%$ \\
\tabincell{c}{`dog',`frog'}& $69.65\%$ & $47.35\%$ & $64.65\%$ & $68.35\%$ & $56.85\%$ & $69.65\%$& $49.90\%$   & $70.25\%$& $71.20\%$ & $\textbf{72.25\%}$  \\
\tabincell{c}{`dog',`truck'}& $80.15\%$ & $52.85\%$ & $78.65\%$ & $77.80\%$  & $70.00\%$ & $79.00\%$& $49.50\%$   & $80.05\%$& $\textbf{80.95\%}$ & $80.40\%$ \\

\end{tabular}
\label{exp3_tbl1_full}

}
\end{table*}

\subsubsection{CMU2008} The second fMRI dataset we consider is CMU2008. It shows the brain activities associated with the meanings of nouns. During the data collection period, the subjects were asked to view $60$ different word-picture from $12$ semantic categories. There are $5$ pictures in each categories and each images is shown to the subject for $6$ times. Therefore, we can get $30$ fMRI images for each semantic category, and each fMRI image is with dimensions $51 \times 61 \times 23$. In this experiment, we consider all the ROIs thus the classified fMRI images are relatively denser than the images we classified in the StarPlus example. Considering the extremely small number of samples in each category, we therefore follow the experiment settings in~\cite{kampa2014sparse} , which combines two similar categories into an integrated class. Specifically, we combine categories $animal$ and $insect$ as class \textbf{Animals}, and categories $tool$ and $furniture$ as class \textbf{Tools}. By doing so, we have $60$ samples in both \textbf{Animals} and \textbf{Tools} classes. We separate the total $120$ fMRI images as training, validation and testing sets, with $50$, $20$ and $50$ images respectively. 

Table~\ref{exp2_tbl2} shows the binary classification results of different models. We notice that SVM can perform classification on this dataset since we include all ROIs, which facilitates the hyperparameter searching procedure. However, its classification accuracies on four subjects are lower than $50\%$. The linear and polynomial model, namely STM, STuM, STTM, and TT classiifer, can only achieve an acceptable performance on a few subjects. Due to the high-dimensional data size, DuSK fails to find a good CP-rank in acceptable time and can not achieve a good classification accuracy. 3D CNN still performs poor due to the very few training samples and high-dimensional feature size. Our proposed two methods still achieve the best classification results on all subjects.

\subsection{CIFAR-10}
    In this experiment, we use the CIFAR-10 dataset~\cite{krizhevsky2009learning} to investigate the fourth claim in Section~\ref{sec:reasonTT}, namely, we can perform different kernel functions on different tensor modes. Here we demonstrate the effect on K-STTM-Prod only. 
    We also randomly select ten class pairs to do binary classification. Without overlap, $50$ samples from the training set of each class are picked randomly for model training and validation respectively, while all the test samples of each class are used for testing. Since each color image is naturally a three-way tensor (pixel-pixel-color), and the first two tensor modes are related to pixel intensity, we therefore utilize the same Gaussian RBF kernel for the first two tensor modes and try a different kernel (linear or polynomial) for the third mode. The parameters $c,d$ in the polynomial kernel $\mat{k}(\mat{x},\mat{y})=(\mat{x}^T\mat{y}+c)^d$ were empirically set to $c=1$ and $d=2$. The baseline case is when the Gaussian RBF kernel is applied to all tensor modes.

Table~\ref{exp3_tbl1_full} lists the classification results. We can observe that K-STTM-Prod still achieves the best accuracy on all class pairs. And by applying a linear or polynomial kernel on the color mode, the classification accuracy of K-STTM-Prod outperforms the baseline case (RBF-RBF-RBF) on nine class pairs, which indicates the potential benefit of employing different kernel functions on different tensor modes when they contain different kind of information. Since the data size of CIFAR-10 is also relatively small, we get a similar observation as the MNIST experiment, namely our method only achieves slightly better classification performance than SVM and 3D CNN on some class pairs. Due to the constrained rank-one model setting, STM can not achieve an acceptable performance. The other two linear classifiers, namely STuM and STTM still perform poor on most of the class pairs. The TT classifier has a similar performance as the STM in this experiment.



\section{CONCLUSIONS AND FUTURE WORKS}
\label{sec:conclusion}
This paper has proposed a tensor train (TT)-based kernel trick for the first time and devised a kernelized support tensor train machine (K-STTM). Assuming a low-rank TT as the prior structure of multi-dimensional data, we first define a corresponding feature mapping scheme that keeps the TT structure in the feature space. Furthermore, two kernel function construction schemes are proposed with consideration of consistency with the TT inner product and the preservation of information, respectively. The feasibility of applying different kernel mappings on the tensor modes with different characteristics is also investigated. Experiments have demonstrated the superiority of K-STTM over conventional approaches for tensorial data in few-sample size problems. 

We further envision two future research directions based on the K-STTM framework. Firstly, instead of constructing a kernel matrix in the K-STTM formula, we will consider building a kernel tensor. We believe that the kernel matrix constructed for each mode can contain different information. Simply multiplying or adding this information may not be the best solution. Subsequently, we propose to stack this information into a 3-way kernel tensor and develop a better way to exploit information in each of the modes. Secondly, we will embed the proposed kernel mapping trick into other kernel-based methods such as LSSVM~\cite{suykens1999least}, kernel PCA~\cite{scholkopf1998nonlinear} etc., such that these methods can directly deal with tensorial data and achieve potentially better performance.

\bibliographystyle{IEEEtran}
\bibliography{IEEEabrv,ijcai18}

\begin{thebibliography}{10}
\providecommand{\url}[1]{#1}
\csname url@rmstyle\endcsname
\providecommand{\newblock}{\relax}
\providecommand{\bibinfo}[2]{#2}
\providecommand\BIBentrySTDinterwordspacing{\spaceskip=0pt\relax}
\providecommand\BIBentryALTinterwordstretchfactor{4}
\providecommand\BIBentryALTinterwordspacing{\spaceskip=\fontdimen2\font plus
\BIBentryALTinterwordstretchfactor\fontdimen3\font minus
  \fontdimen4\font\relax}
\providecommand\BIBforeignlanguage[2]{{%
\expandafter\ifx\csname l@#1\endcsname\relax
\typeout{** WARNING: IEEEtran.bst: No hyphenation pattern has been}%
\typeout{** loaded for the language `#1'. Using the pattern for}%
\typeout{** the default language instead.}%
\else
\language=\csname l@#1\endcsname
\fi
#2}}

\bibitem{dai2006tensor}
G.~Dai and D.-Y. Yeung, ``Tensor embedding methods,'' in \emph{AAAI}, vol.~6,
  2006, pp. 330--335.

\bibitem{MPCA}
H.~Lu, K.~N. Plataniotis, and A.~N. Venetsanopoulos, ``Mpca: Multilinear
  principal component analysis of tensor objects,'' \emph{IEEE transactions on
  Neural Networks}, vol.~19, no.~1, pp. 18--39, 2008.

\bibitem{boser1992training}
B.~E. Boser, I.~M. Guyon, and V.~N. Vapnik, ``A training algorithm for optimal
  margin classifiers,'' in \emph{Proceedings of the fifth annual workshop on
  Computational learning theory}.\hskip 1em plus 0.5em minus 0.4em\relax ACM,
  1992, pp. 144--152.

\bibitem{STM}
D.~Tao, X.~Li, W.~Hu, S.~Maybank, and X.~Wu, ``Supervised tensor learning,'' in
  \emph{Fifth IEEE International Conference on Data Mining (ICDM'05)}.\hskip
  1em plus 0.5em minus 0.4em\relax IEEE, 2005, pp. 8--pp.

\bibitem{nguyen2015tensor}
T.~D. Nguyen, T.~Tran, D.~Phung, and S.~Venkatesh, ``Tensor-variate restricted
  boltzmann machines,'' in \emph{Twenty-Ninth AAAI Conference on Artificial
  Intelligence}, 2015.

\bibitem{phan2010tensor}
A.~H. Phan and A.~Cichocki, ``Tensor decompositions for feature extraction and
  classification of high dimensional datasets,'' \emph{Nonlinear theory and its
  applications, IEICE}, vol.~1, no.~1, pp. 37--68, 2010.

\bibitem{liu2013tensor}
J.~Liu, P.~Musialski, P.~Wonka, and J.~Ye, ``Tensor completion for estimating
  missing values in visual data,'' \emph{IEEE transactions on pattern analysis
  and machine intelligence}, vol.~35, no.~1, pp. 208--220, 2013.

\bibitem{fanaee2016tensor}
H.~Fanaee-T and J.~Gama, ``Tensor-based anomaly detection: An interdisciplinary
  survey,'' \emph{Knowledge-Based Systems}, vol.~98, pp. 130--147, 2016.

\bibitem{guo2012tensor}
W.~Guo, I.~Kotsia, and I.~Patras, ``Tensor learning for regression,''
  \emph{IEEE Transactions on Image Processing}, vol.~21, no.~2, pp. 816--827,
  2012.

\bibitem{lebedev2014speeding}
V.~Lebedev, Y.~Ganin, M.~Rakhuba, I.~Oseledets, and V.~Lempitsky, ``Speeding-up
  convolutional neural networks using fine-tuned cp-decomposition,''
  \emph{arXiv preprint arXiv:1412.6553}, 2014.

\bibitem{novikov2015tensorizing}
A.~Novikov, D.~Podoprikhin, A.~Osokin, and D.~P. Vetrov, ``Tensorizing neural
  networks,'' in \emph{Advances in Neural Information Processing Systems},
  2015, pp. 442--450.

\bibitem{oseledets2011tensor}
I.~V. Oseledets, ``{Tensor-train decomposition},'' \emph{SIAM Journal on
  Scientific Computing}, vol.~33, no.~5, pp. 2295--2317, 2011.

\bibitem{signoretto2011kernel}
M.~Signoretto, L.~De~Lathauwer, and J.~A. Suykens, ``A kernel-based framework
  to tensorial data analysis,'' \emph{Neural networks}, vol.~24, no.~8, pp.
  861--874, 2011.

\bibitem{zhao2013kernel}
Q.~Zhao, G.~Zhou, T.~Adal{\i}, L.~Zhang, and A.~Cichocki, ``Kernel-based tensor
  partial least squares for reconstruction of limb movements,'' in \emph{2013
  IEEE International Conference on Acoustics, Speech and Signal
  Processing}.\hskip 1em plus 0.5em minus 0.4em\relax IEEE, 2013, pp.
  3577--3581.

\bibitem{kotsia2012higher}
I.~Kotsia, W.~Guo, and I.~Patras, ``Higher rank support tensor machines for
  visual recognition,'' \emph{Pattern Recognition}, vol.~45, no.~12, pp.
  4192--4203, 2012.

\bibitem{kotsia2011support}
I.~Kotsia and I.~Patras, ``Support tucker machines,'' in \emph{CVPR
  2011}.\hskip 1em plus 0.5em minus 0.4em\relax IEEE, 2011, pp. 633--640.

\bibitem{linearSTTM}
\BIBentryALTinterwordspacing
C.~Chen, K.~Batselier, C.~Y. Ko, and N.~Wong, ``A support tensor train
  machine,'' \emph{CoRR}, vol. abs/1804.06114, 2018. [Online]. Available:
  \url{http://arxiv.org/abs/1804.06114}
\BIBentrySTDinterwordspacing

\bibitem{he2014dusk}
L.~He, X.~Kong, P.~S. Yu, X.~Yang, A.~B. Ragin, and Z.~Hao, ``Dusk: A dual
  structure-preserving kernel for supervised tensor learning with applications
  to neuroimages,'' in \emph{Proceedings of the 2014 SIAM International
  Conference on Data Mining}.\hskip 1em plus 0.5em minus 0.4em\relax SIAM,
  2014, pp. 127--135.

\bibitem{he2017multi}
L.~He, C.-T. Lu, H.~Ding, S.~Wang, L.~Shen, P.~S. Yu, and A.~B. Ragin,
  ``Multi-way multi-level kernel modeling for neuroimaging classification,'' in
  \emph{Proceedings of the IEEE Conference on Computer Vision and Pattern
  Recognition}, 2017, pp. 356--364.

\bibitem{he2017kernelized}
L.~He, C.-T. Lu, G.~Ma, S.~Wang, L.~Shen, P.~S. Yu, and A.~B. Ragin,
  ``Kernelized support tensor machines,'' in \emph{Proceedings of the 34th
  International Conference on Machine Learning-Volume 70}.\hskip 1em plus 0.5em
  minus 0.4em\relax JMLR. org, 2017, pp. 1442--1451.

\bibitem{orus2014practical}
R.~Or{\'u}s, ``A practical introduction to tensor networks: Matrix product
  states and projected entangled pair states,'' \emph{Annals of Physics}, vol.
  349, pp. 117--158, 2014.

\bibitem{tensormultiview}
L.~Houthuys and J.~A.~K.~Suykens, \emph{Tensor Learning in Multi-view Kernel
  PCA: 27th International Conference on Artificial Neural Networks, Rhodes,
  Greece, October 4-7, 2018, Proceedings, Part II}, 09 2018, pp. 205--215.

\bibitem{schur1911bemerkungen}
J.~Schur, ``Bemerkungen zur theorie der beschr{\"a}nkten bilinearformen mit
  unendlich vielen ver{\"a}nderlichen.'' \emph{Journal f{\"u}r die reine und
  Angewandte Mathematik}, vol. 140, pp. 1--28, 1911.

\bibitem{gupta2013natural}
A.~Gupta, M.~Ayhan, and A.~Maida, ``Natural image bases to represent
  neuroimaging data,'' in \emph{International conference on machine learning},
  2013, pp. 987--994.

\bibitem{chen2017parallelized}
Z.~Chen, K.~Batselier, J.~A. Suykens, and N.~Wong, ``Parallelized tensor train
  learning of polynomial classifiers,'' \emph{IEEE Transactions on Neural
  Networks and Learning Systems}, 2017.

\bibitem{lecun-mnisthandwrittendigit-2010}
\BIBentryALTinterwordspacing
Y.~LeCun and C.~Cortes, ``{MNIST} handwritten digit database,'' 2010. [Online].
  Available: \url{http://yann.lecun.com/exdb/mnist/}
\BIBentrySTDinterwordspacing

\bibitem{mitchell2008predicting}
T.~M. Mitchell, S.~V. Shinkareva, A.~Carlson, K.-M. Chang, V.~L. Malave, R.~A.
  Mason, and M.~A. Just, ``Predicting human brain activity associated with the
  meanings of nouns,'' \emph{science}, vol. 320, no. 5880, pp. 1191--1195,
  2008.

\bibitem{kampa2014sparse}
K.~Kampa, S.~Mehta, C.-A. Chou, W.~A. Chaovalitwongse, and T.~J. Grabowski,
  ``Sparse optimization in feature selection: application in neuroimaging,''
  \emph{Journal of Global Optimization}, vol.~59, no. 2-3, pp. 439--457, 2014.

\bibitem{krizhevsky2009learning}
A.~Krizhevsky and G.~Hinton, ``Learning multiple layers of features from tiny
  images,'' Citeseer, Tech. Rep., 2009.

\bibitem{suykens1999least}
J.~A. Suykens and J.~Vandewalle, ``Least squares support vector machine
  classifiers,'' \emph{Neural processing letters}, vol.~9, no.~3, pp. 293--300,
  1999.

\bibitem{scholkopf1998nonlinear}
B.~Sch{\"o}lkopf, A.~Smola, and K.-R. M{\"u}ller, ``Nonlinear component
  analysis as a kernel eigenvalue problem,'' \emph{Neural computation},
  vol.~10, no.~5, pp. 1299--1319, 1998.

\end{thebibliography}

\newpage
\appendix
We list the details of the experimental settings for each method over all experiments. Hyperparameters were determined via a grid search, for which we provide the search range.

Table~\ref{app_table1} shows the hyperparameters search range of different methods in MNIST datasets. We provide a wide search range of performance trade-off parameter $C$ and Gaussian kernel width $\sigma$ such that it can fulfill the requirement for all methods. Though STM is a tensor-based method, it assume its weight tensor to be rank-one, thus there is no need to determine the tensor rank. We mention that the upper bound of tensor rank in STuM, STTM and K-STTM is $28$.
STuM and the polynomial TT classifier assume the same rank over all tensor modes, namely $R_1 = R_2=  \cdots = R_d = R$. For the meaning of other hyperparameters mentioned in Table~\ref{app_table1}, we refer the reader to the corresponding papers for more information.

Table~\ref{app_table2} and~\ref{app_table3} show the parameter search range in the StarPlus and CMU2008 fMRI datasets respectively. Due to the high-dimensional fMRI image size, we increased the tensor rank search range for the tensor-based methods. STuM assumes that the upper bound of the Tucker rank is limited to the smallest dimension of the two fMRI datasets, namely 8 and 23 respectively. Thus the tensor rank search range of STuM is relatively smaller than other methods. 
As for the polynomial TT classifier, although it is based on the scalable TT structure, we could not further increase the parameter search range since this increases the memory consumption a lot and made the training speed extremely slow.   

Table~\ref{app_table4} lists the search range for the CIFAR-10 dataset. Since the dimension of the third mode of a color image is only $3$, we therefore do not compress this mode and keep the TT-rank $R_3$ at $3$ in both the STTM and K-STTM methods.

\begin{table}[H]
\scriptsize
\newcommand{\tabincell}[2]{\begin{tabular}{@{}#1@{}}#2\end{tabular}}
\centering
\caption{\label{cmu2008comparison} Hyperparameters search range of different methods in MNIST datasets.}
\vspace{1ex}
\begin{tabular}{@{}ll@{}}

Methods & \tabincell{c}{~~~~~~~~~~~~~~~~~Hyperparameters search range ~~~~~~~~~~~~~~~} \\
\hline
SVM & \tabincell{l}{performance trade-off parameter $C$: \{$10^{-6}$, $10^{-5}$, \ldots, $10^{8}$, $10^{9}$ \} \\ Gaussian kernel width parameter $\sigma$: \{$10^{-6}$, $10^{-5}$, \ldots, $10^{8}$, $10^{9}$ \}}\\
\hline
STM & \tabincell{l}{performance trade-off parameter $C$: \{$10^{-6}$, $10^{-5}$, \ldots, $10^{8}$, $10^{9}$ \}}\\
\hline
STuM & \tabincell{l}{performance trade-off parameter $C$: \{$10^{-6}$, $10^{-5}$, \ldots, $10^{8}$, $10^{9}$ \}\\
Tucker rank $R$: \{$2$, $3$, $\ldots$, $28$\} \quad \quad \\
threshold criterion $\epsilon$: \{$10^{-6}$, $10^{-5}$, \ldots, $10^{-1}$, $10^{0}$ \} \\
  number of iterations: \{$1$, $2$, $3$, $4$, $5$\}\\
}\\
\hline
STTM & \tabincell{l}{tensor train ranks $R_2$: \{$2$, $3$, $\ldots$, $28$\} \\
relative error $\epsilon$ of TT approximation of training data: $10^{-5}$}\\
\hline
DuSK & \tabincell{l}{performance trade-off parameter $C$: \{$10^{-6}$, $10^{-5}$, \ldots, $10^{8}$, $10^{9}$ \} \\ Gaussian kernel width parameter $\sigma$: \{$10^{-6}$, $10^{-5}$, \ldots, $10^{8}$, $10^{9}$ \}\\
CP ranks $R$: \{$2$, $3$, $\ldots$, $40$\}}\\
\hline
3D CNN & \tabincell{l}{hidden layer size: \{$10$, $30$, $50$, $100$, $150$, $200$\}}\\
\hline
TT classifier & \tabincell{l}{Tensor train order $d$: \{$15$, $20$, $25$, $30$, $35$, $40$\}\\
Tensor train dimensions $n$: \{$2$, $3$, \ldots, $10$ \} \\ Tensor train ranks $R$: \{$2$, $3$, \ldots, $20$ \}}\\
\hline
\tabincell{l}{K-STTM-\\Prod/Sum} & \tabincell{l}{performance trade-off parameter $C$: \{$10^{-6}$, $10^{-5}$, \ldots, $10^{8}$, $10^{9}$ \} \\ Gaussian kernel width parameter $\sigma$: \{$10^{-6}$, $10^{-5}$, \ldots, $10^{8}$, $10^{9}$ \} \\ Tensor train ranks $R_2$: \{$2$, $3$, \ldots, $28$ \}}\\
\hline
\end{tabular}
\label{app_table1}
\end{table}

\begin{table}[H]
\scriptsize
\newcommand{\tabincell}[2]{\begin{tabular}{@{}#1@{}}#2\end{tabular}}
\centering
\caption{\label{cmu2008comparison} Hyperparameters search range of different methods in StarPlus fMRI datasets.}
\vspace{1ex}
\begin{tabular}{@{}ll@{}}

Methods & \tabincell{c}{~~~~~~~~~~~~~~~~~Hyperparameters search range ~~~~~~~~~~~~~~~} \\
\hline
SVM & \tabincell{l}{performance trade-off parameter $C$: \{$10^{-6}$, $10^{-5}$, \ldots, $10^{8}$, $10^{9}$ \} \\ Gaussian kernel width parameter $\sigma$: \{$10^{-6}$, $10^{-5}$, \ldots, $10^{8}$, $10^{9}$ \}}\\
\hline
STM & \tabincell{l}{performance trade-off parameter $C$: \{$10^{-6}$, $10^{-5}$, \ldots, $10^{8}$, $10^{9}$ \}}\\
\hline
STuM & \tabincell{l}{performance trade-off parameter $C$: \{$10^{-6}$, $10^{-5}$, \ldots, $10^{8}$, $10^{9}$ \}\\
Tucker rank $R$: \{$2$, $3$, $\ldots$, $8$\} \quad \quad \\ threshold criterion $\epsilon$: \{$10^{-6}$, $10^{-5}$, \ldots, $10^{-1}$, $10^{0}$ \} \\ number of iterations: \{$1$, $2$, $3$, $4$, $5$\}}\\
\hline
STTM & \tabincell{l}{tensor train ranks $R_2$: \{$60$, $80$, $100$, $\ldots$, $200$\}; $R_3$: \{$2$, $4$, $6$, $8$\}\\
relative error $\epsilon$ of TT approximation of training data: $10^{-5}$}\\
\hline
DuSK & \tabincell{l}{performance trade-off parameter $C$: \{$10^{-6}$, $10^{-5}$, \ldots, $10^{8}$, $10^{9}$ \} \\ Gaussian kernel width parameter $\sigma$: \{$10^{-6}$, $10^{-5}$, \ldots, $10^{8}$, $10^{9}$ \}\\
CP ranks $R$: \{$60$, $80$, $100$, $\ldots$, $180$, $200$\}}\\
\hline
3D CNN & \tabincell{l}{hidden layer size: \{$10$, $30$, $50$, $100$, $150$, $200$\}}\\
\hline
TT classifier & \tabincell{l}{Tensor train order $d$: \{$15$, $20$, $25$, $30$, $35$, $40$\}\\
Tensor train dimensions $n$: \{$2$, $3$, \ldots, $10$ \} \\ Tensor train ranks $R$: \{$2$, $3$, \ldots, $20$ \}}\\
\hline
\tabincell{l}{K-STTM-\\Prod/Sum} & \tabincell{l}{performance trade-off parameter $C$: \{$10^{-6}$, $10^{-5}$, \ldots, $10^{8}$, $10^{9}$ \} \\ Gaussian kernel width parameter $\sigma$: \{$10^{-6}$, $10^{-5}$, \ldots, $10^{8}$, $10^{9}$ \} \\ Tensor train ranks $R_2$: \{$60$, $80$, $100$, $\ldots$, $200$\}; $R_3$: \{$2$, $4$, $6$, $8$\}}\\
\hline
\end{tabular}
\label{app_table2}
\end{table}

\begin{table}[H]
\scriptsize
\newcommand{\tabincell}[2]{\begin{tabular}{@{}#1@{}}#2\end{tabular}}
\centering
\caption{Hyperparameters search range of different methods in CMU2008 fMRI datasets.}
\vspace{1ex}
\begin{tabular}{@{}ll@{}}

Methods & \tabincell{c}{~~~~~~~~~~~~~~~~~Hyperparameters search range ~~~~~~~~~~~~~~~} \\
\hline
SVM & \tabincell{l}{performance trade-off parameter $C$: \{$10^{-6}$, $10^{-5}$, \ldots, $10^{8}$, $10^{9}$ \} \\ Gaussian kernel width parameter $\sigma$: \{$10^{-6}$, $10^{-5}$, \ldots, $10^{8}$, $10^{9}$ \}}\\
\hline
STM & \tabincell{l}{performance trade-off parameter $C$: \{$10^{-6}$, $10^{-5}$, \ldots, $10^{8}$, $10^{9}$ \}}\\
\hline
STuM & \tabincell{l}{performance trade-off parameter $C$: \{$10^{-6}$, $10^{-5}$, \ldots, $10^{8}$, $10^{9}$ \}\\
Tucker rank $R$: \{$2$, $3$, $\ldots$, $23$\} \quad \quad \\ threshold criterion $\epsilon$: \{$10^{-6}$, $10^{-5}$, \ldots, $10^{-1}$, $10^{0}$ \} \\ number of iterations: \{$1$, $2$, $3$, $4$, $5$\}}\\
\hline
STTM & \tabincell{l}{tensor train ranks $R_2$: \{$10$, $20$, $\ldots$, $100$\}; $R_3$: \{$2$, $4$, $\ldots$, $20$\} \\
relative error $\epsilon$ of TT approximation of training data: $10^{-5}$}\\
\hline
DuSK & \tabincell{l}{performance trade-off parameter $C$: \{$10^{-6}$, $10^{-5}$, \ldots, $10^{8}$, $10^{9}$ \} \\ Gaussian kernel width parameter $\sigma$: \{$10^{-6}$, $10^{-5}$, \ldots, $10^{8}$, $10^{9}$ \}\\
CP ranks $R$: \{$10$, $20$, $\ldots$, $90$, $100$\}}\\
\hline
3D CNN & \tabincell{l}{hidden layer size: \{$10$, $30$, $50$, $100$, $150$, $200$\}}\\
\hline
TT classifier & \tabincell{l}{Tensor train order $d$: \{$15$, $20$, $25$, $30$, $35$, $40$\}\\
Tensor train dimensions $n$: \{$2$, $3$, \ldots, $10$ \} \\ Tensor train ranks $R$: \{$2$, $3$, \ldots, $20$ \}}\\
\hline
\tabincell{l}{K-STTM-\\Prod/Sum} & \tabincell{l}{performance trade-off parameter $C$: \{$10^{-6}$, $10^{-5}$, \ldots, $10^{8}$, $10^{9}$ \} \\ Gaussian kernel width parameter $\sigma$: \{$10^{-6}$, $10^{-5}$, \ldots, $10^{8}$, $10^{9}$ \} \\ Tensor train ranks $R_2$: \{$10$, $20$, $\ldots$, $100$\}; $R_3$: \{$2$, $4$, $\ldots$, $20$\}}\\
\hline
\end{tabular}
\label{app_table3}
\end{table}

\begin{table}[H]
\scriptsize
\newcommand{\tabincell}[2]{\begin{tabular}{@{}#1@{}}#2\end{tabular}}
\centering
\caption{Hyperparameters search range of different methods in CIFAR-10 datasets.}
\vspace{1ex}
\begin{tabular}{@{}ll@{}}

Methods & \tabincell{c}{~~~~~~~~~~~~~~~~~Hyperparameters search range ~~~~~~~~~~~~~~~} \\
\hline
SVM & \tabincell{l}{performance trade-off parameter $C$: \{$10^{-6}$, $10^{-5}$, \ldots, $10^{8}$, $10^{9}$ \} \\ Gaussian kernel width parameter $\sigma$: \{$10^{-6}$, $10^{-5}$, \ldots, $10^{8}$, $10^{9}$ \}}\\
\hline
STM & \tabincell{l}{performance trade-off parameter $C$: \{$10^{-6}$, $10^{-5}$, \ldots, $10^{8}$, $10^{9}$ \}}\\
\hline
STuM & \tabincell{l}{performance trade-off parameter $C$: \{$10^{-6}$, $10^{-5}$, \ldots, $10^{8}$, $10^{9}$ \}\\
Tucker rank $R$: \{$2$, $3$\} \quad \quad \\ 
threshold criterion $\epsilon$: \{$10^{-6}$, $10^{-5}$, \ldots, $10^{-1}$, $10^{0}$ \} \\ 
number of iterations: \{$1$, $2$, $3$, $4$, $5$\}}\\
\hline
STTM & \tabincell{l}{tensor train ranks $R_2$: \{$2$, $4$, $\ldots$, $30$, $32$\}; $R_3$: \{$3$\} \\
relative error $\epsilon$ of TT approximation of training data: $10^{-5}$}\\
\hline
DuSK & \tabincell{l}{performance trade-off parameter $C$: \{$10^{-6}$, $10^{-5}$, \ldots, $10^{8}$, $10^{9}$ \} \\ Gaussian kernel width parameter $\sigma$: \{$10^{-6}$, $10^{-5}$, \ldots, $10^{8}$, $10^{9}$ \}\\
CP ranks $R$: \{$2$, $4$, $\ldots$, $48$, $50$\}}\\
\hline
3D CNN & \tabincell{l}{hidden layer size: \{$10$, $30$, $50$, $100$, $150$, $200$\}}\\
\hline
TT classifier & \tabincell{l}{Tensor train order $d$: \{$15$, $20$, $25$, $30$, $35$, $40$\}\\
Tensor train dimensions $n$: \{$2$, $3$, \ldots, $10$ \} \\ Tensor train ranks $R$: \{$2$, $3$, \ldots, $20$\}}\\
\hline
\tabincell{l}{K-STTM-\\Prod} & \tabincell{l}{performance trade-off parameter $C$: \{$10^{-6}$, $10^{-5}$, \ldots, $10^{8}$, $10^{9}$ \} \\ Gaussian kernel width parameter $\sigma$: \{$10^{-6}$, $10^{-5}$, \ldots, $10^{8}$, $10^{9}$ \} \\ Tensor train ranks $R_2$: \{$2$, $4$, $\ldots$, $30$, $32$\}; $R_3$: \{$3$\}}\\
\hline
\end{tabular}
\label{app_table4}
\end{table}

\end{document}